\documentclass[letterpaper, 10 pt, conference]{ieeeconf}  %

\IEEEoverridecommandlockouts                              %

\usepackage{graphicx} 

\makeatletter
\let\MYcaption\@makecaption
\makeatother

\usepackage[font=footnotesize]{subcaption}

\makeatletter
\let\@makecaption\MYcaption
\makeatother

\usepackage{textcomp}
\usepackage{array}
\usepackage{booktabs}
\usepackage[dvipsnames]{xcolor}
\usepackage{tikz}
\usepackage[export]{adjustbox}
\usepackage{url}
\usepackage{multirow}
\usepackage{soul}
\usepackage{amsmath, amssymb}
\usepackage{amsfonts}
\usepackage{systeme}
\usepackage{algorithm}
\usepackage[noend]{algorithmic}
\usepackage{comment}
\usepackage{hyperref}
\usepackage{physics}
\usepackage{multirow}
\usepackage{mwe}
\usepackage[percent]{overpic}

\usepackage[style=ieee,mincitenames=1,maxcitenames=2,bibencoding=utf8,url=false]{biblatex}
\renewcommand*{\bibfont}{\footnotesize}
\AtEveryBibitem{\clearfield{pages}
 \clearfield{volume}%
  \clearfield{number}
  \clearfield{month}}

\usepackage[binary-units=true]{siunitx}
\usepackage{physics}

\usepackage[percent]{overpic}
\usepackage{tabularx}
\usepackage[export]{adjustbox}

\long\def\invis#1{}
\newcommand\sect[1]{Section~\ref{#1}}
\newcommand\tab[1]{Table~\ref{#1}}
\newcommand\fig[1]{Fig.~\ref{#1}}
\newcommand\eq[1]{Equation~(\ref{#1})}

\DeclareMathOperator*{\argmin}{arg\,min}

\DeclareSIUnit\rpm{RPM}

\newcommand\revisedcom[1]{\textcolor{black}{#1}}
\newcommand\todo[1]{\textcolor{blue}{todo: #1}}

\newcommand\catabot{\emph{Catabot}}
\long\def\invis#1{}

\DeclareMathOperator{\atantwo}{atan2}

\title{\LARGE \bf
Active Learning-augmented Intention-aware Obstacle Avoidance of Autonomous Surface Vehicles in High-traffic Waters
}

\author{Mingi Jeong$^1$, Arihant Chadda$^2$, and Alberto Quattrini Li$^1$%
\thanks{$^1$Computer Science Department, Dartmouth College, Hanover, NH USA {\tt\footnotesize \{mingi.jeong.gr, alberto.quattrini.li\}@dartmouth.edu.}}
\thanks{$^2$IQT Labs, Tysons, VA USA {\tt\footnotesize achadda@iqt.org.}}
\thanks{We would like to thank Monika Roznere and Sam Lensgraf for help with field experiments and McGill Bellairs Research Institute for experimental sites. This work is supported in part by the Burke Research Initiation Award, NSF CNS-1919647, 2144624, OIA1923004, and NH Sea Grant.}%
}

\begin{document}

\maketitle
\thispagestyle{empty}
\pagestyle{empty}

\begin{abstract}
This paper enhances the obstacle avoidance of Autonomous Surface Vehicles (ASVs) for safe navigation in high-traffic waters with an active state estimation of obstacle's passing intention and reducing its uncertainty. We introduce a topological modeling of passing intention of obstacles, which can be applied to varying encounter situations based on the inherent embedding of topological concepts in COLREGs. With a Long Short-Term Memory (LSTM) neural network, we classify the passing intention of obstacles. Then, for determining the ASV maneuver, we propose a multi-objective optimization framework including information gain about the passing obstacle intention and safety. We validate the proposed approach under extensive Monte Carlo simulations (\num[group-minimum-digits={3},group-separator={,}]{2400} runs) with a varying number of obstacles, dynamic properties, encounter situations, and different behavioral patterns of obstacles (cooperative, non-cooperative). We also present the results from a real marine accident case study as well as real-world experiments of a real ASV with environmental disturbances, showing successful collision avoidance with our strategy in real-time.
\end{abstract}

\section{Introduction}

This paper demonstrates best-in-class navigation safety for Autonomous Surface Vehicles (ASVs) in high-traffic waterways through a novel approach to understanding the intentions of passing vehicles for obstacle avoidance. Generally, the intentions of passing vehicles are not known by the ego ASV, as marine vessels do not share their intentions with others. This lack of knowledge, together with the absence of clearly marked lanes as on roads, among other challenges, makes navigation extremely difficult, resulting in potentially risky situations. This challenge is recognized to significantly hinder the advancement of ASVs' autonomy and their use in high-impact applications, including environmental monitoring and shipping~\cite{UN-2022}.  To address this challenge and enable safe navigation, the core element of our proposed approach involves topological modeling of obstacle passing, learning-based \revisedcom{passing intention classification}, and \revisedcom{then taking active intention-aware actions} that reduce the associated uncertainty of passing -- see \fig{fig:beauty} for a visual explanation.

\revisedcom{There are many works on intention prediction of other vehicles, e.g., self-driving cars in urban environments \cite{vectornet-2020, intention-aware-dynamicattention-2022, Li_Zhao_Xu_Wang_Chen_Dai_2021-lane-change-intention}. However,  %
these methods are not directly applicable in the maritime domain, because of the unfavorable characteristics in aquatic environments.} \revisedcom{Specifically, aquatic} environments are characterized by their \textbf{(1)} unstructured and open nature compared to ground vehicles operating on clearly marked roads. Additionally, \revisedcom{ASVs}  \textbf{(2)} suffer from \revisedcom{unpredictable roll, pitch, and yaw changes leading to noisy sensor measurements and limited maneuverability due to water dynamics, which underscores the importance of accurate intention prediction.}
While there are established traffic rules, specifically the International Regulations for Preventing Collisions at Sea (COLREGs) \cite{colreg}, which regulate evasive maneuvers in various scenarios, the rules introduce \textbf{(3)} some ambiguity. For instance, terms like `large enough to be readily apparent to another vessel' and `head-on situation where two vessels meet on a nearly reciprocal course' lack explicit definitions, leading to varying levels of compliance and inconsistent interpretation when in their operation.
\invis{as maritime vessels may be unable to react quickly enough to avoid collision if the intention is misinterpreted.}
\invis{and noisy sensor measurements due to water dynamics, and have limited maneuverability to react in case of misinterpretation of the intent.} 
\invis{
\begin{figure}[t!]
    \centering
    \includegraphics[width=0.8\columnwidth]{figs/fig1-beauty-horizontal-new-new.pdf}
    \caption{Active learning-augmented intention-aware obstacle avoidance in multiple encounters. With an uncertain scenario where obstacles are passing with respect to the ego ASV (top), the ego ASV classifies the topological passing side and actively determines an action to increase information gain (bottom), not passively waiting to find the passing intent. \todo{figure change to compare with non-intention}}
    \label{fig:beauty}
    \vspace{-2em}
\end{figure}
}

\begin{figure}[tbp]
    \centering
    \begin{minipage}{0.7\columnwidth}
        \centering
        \begin{subfigure}[b]{1\linewidth}
            \includegraphics[width=\textwidth]{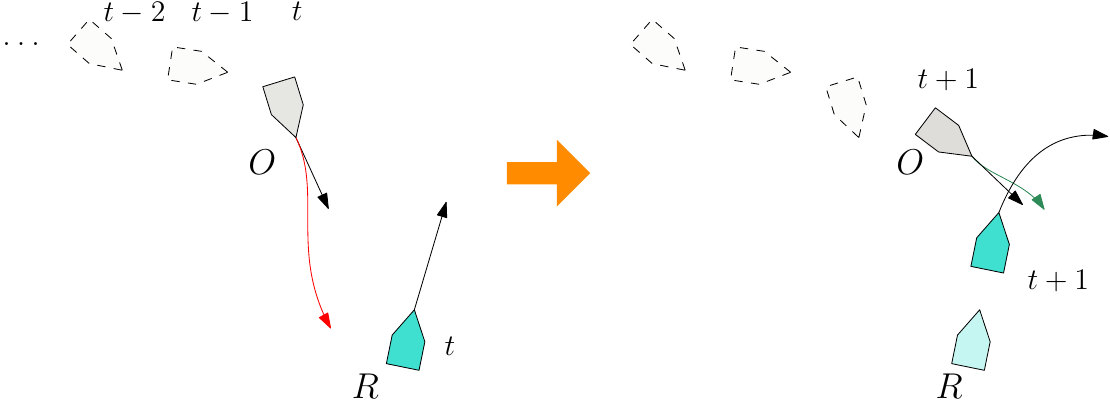}
            \vspace{-0.5cm}
            \caption{State-of-the-art method without intention-awareness}
            \label{fig:beauty-literature}
        \end{subfigure}
    \end{minipage}
    \hrule
    \begin{minipage}{0.75\columnwidth}
        \centering
        \begin{subfigure}[b]{1\linewidth} 
            \includegraphics[width=\textwidth]{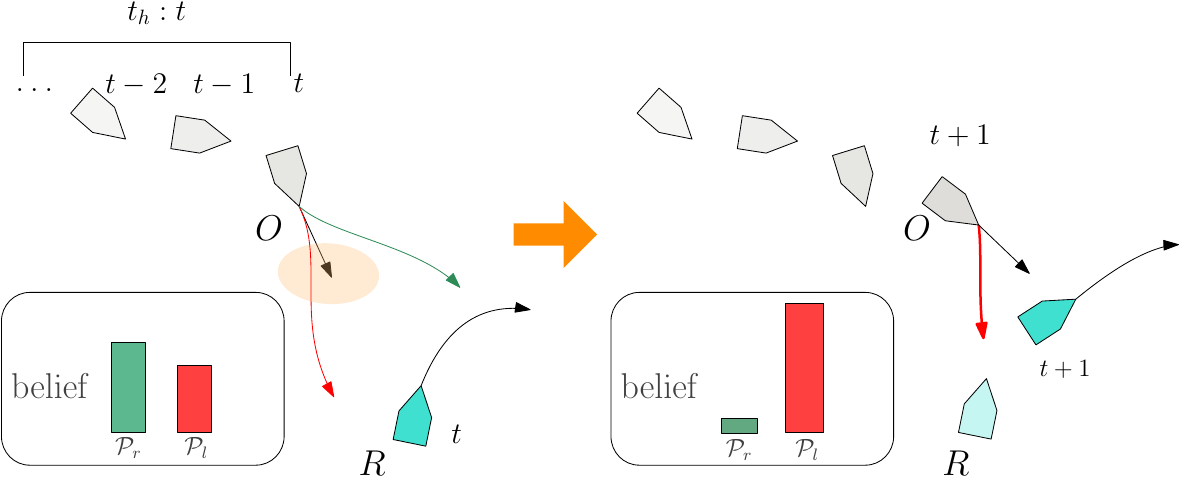}
            \vspace{-0.5cm}
            \caption{Proposed method with intention-awareness}
            \label{fig:beauty-ours}
        \end{subfigure}
    \end{minipage}
    \vspace{-0.15cm}
    \caption{\revisedcom{From time $t$ to $t+1$, controlled ASV, $R$'s collision avoidance behavior by state-of-the-art method vs.\ proposed method using active learning-augmented intention-awareness under an uncertain scenario where an obstacle, $O$ approaches from the left side of $R$: (\textbf{a}) At $t$, the state-of-the-art, lacking intention-awareness, predicts that O will pass on the left side (red) of $R$ and thus $R$ maintains its course as a \textit{stand-on} vessel. At $t+1$, $R$ realizes $O$ is attempting to pass on the right side (green) of $R$, resulting in a nearmiss. $R$ did a hard turn-over but it is too late. (\textbf{b}) At $t$, our method classifies the topological passing side based on the historical data from $t_h$ to $t$ and actively determines an action to increase information gain, i.e., to decrease the probability of passing on the right side (green), which is risky due to bow crossing. This proactive action with \textit{good seamanship} despite a \textit{stand-on} status leads to a safe clearance at $t+1$.}}
    \vspace{-0.75cm}
    \label{fig:beauty}
\end{figure}

Many \revisedcom{prior} works on intention inference in the maritime domain \revisedcom{primarily} focused on trajectory prediction \cite{RNN-vessel-trajectory-prediction-2021, Sharif-prediction-JON-2021, Li_Jiao_Yang_-trajectory-2023}. However, the predicted trajectories show significant errors (discrepancies on the order of hundreds of meters along the prediction horizon \cite{RNN-vessel-trajectory-prediction-2021, Li_Jiao_Yang_-trajectory-2023}). Such errors can be acceptable while navigating in the \revisedcom{open} ocean, but they are not in congested areas. \invis{In contrast to the centralized air traffic control, different from other traffics, e.g., centralized air traffic control.}\revisedcom{Also, maritime collision avoidance is handled by each vehicle, not by a centralized traffic control. Therefore, an ASV is required to perform semantic inference from its ego-centric perspective.} 

While higher-level maneuver intention approaches, like inferring rule violations by obstacles \cite{Leavitt-MIT-thesis, KAIST-intent-avoidance2022} have been introduced, \revisedcom{there are some limitations affecting the overall navigation safety:} vehicles tend to take passive actions \revisedcom{according to occurrences of the obstacles' rule violations, rather than preemptive actions. Moreover, these works assume homogeneous traffic behaviors (same as an ego-vehicle and same across obstacles), which is unrealistic.} In practical situations, proactive \revisedcom{and large} actions are essential to reduce uncertainty and mitigate risks associated with hidden intentions \revisedcom{to align} with \revisedcom{key} principles such as COLREGs \cite{colreg, García-Maza_Argüelles-colreg-2022}.

To address the challenges outlined above, we present an innovative approach termed \textbf{active learning-augmented intention-aware obstacle avoidance} designed for handling \revisedcom{single- or multi-obstacle} encounters, without \revisedcom{the ego ASV's} explicit communication as to other vehicles' intentions. \revisedcom{Thus, the proposed approach aligns} with the fundamental principle of \revisedcom{the maritime convention, i.e., \textit{proactive action}, denoted in} \cite{colreg}, enabling the ego-vehicle to exhibit \textit{good seamanship} \revisedcom{and avoid risky situations}, even in \textit{stand-on} status. Specifically, the main contributions of this paper are: %
\begin{itemize}
    \invis{
    \item \revisedcom{novel decision-making that optimizes passing intent, improving the safety of maritime navigation to guide ego vehicle actions in complex scenarios;}
    }
    \item topological modeling of passing based on \revisedcom{maritime navigation}'s inherent \revisedcom{conceptual topology} and \revisedcom{implementation of LSTM-backbone-based intention classification};
    \item a novel multi-objective local planner that includes an active strategy to increase information gain in uncertain encounters about the passing intention of obstacles, while ensuring collision avoidance; and
    \item implementation in Robot Operating System (ROS) with comprehensive analysis through extensive Monte Carlo simulations, experiments in the ocean with a real ASV, and a real-world accident case study successfully demonstrating safe and real-time collision avoidance. 
\end{itemize}

This work represents a first effort to include in collision avoidance strategies the reduction of uncertainty regarding the intention of other obstacles, with the overall goal to improve the ASV navigation safety.

\section{Related Work}\label{sec:background}

Several methods appeared in the literature for obstacle avoidance, primarily for ground robots~\cite{Alonse-Mora_2021-where-to-go, intention-aware-rus-2013, intention-aware-dynamicattention-2022, mcts-interpretable-2021}, and some in the maritime domain~\cite{VO-COLREG-kuwata, KAIST-intent-avoidance2022, Wang_2020-colreg-inference}. Here we focus on those methods that aim to predict the intentions of other vehicles, given that such information is fundamental for safe navigation as discussed in the previous section.

Many studies employ intention-awareness primarily through predictions of vessel trajectory, including \revisedcom{learning-based} approaches like Recurrent Neural Networks (RNNs) \cite{RNN-vessel-trajectory-prediction-2021}, Variational RNN \cite{Nguyen-2018-VRNN},  Bayesian \revisedcom{modeling} based on a Gaussian Process \cite{Rong-2019-gaussian-ship-trajectory},\invis{\revisedcom{multi-layer perceptrons} \cite{Valsamis-2017-trajectory-ML},} and Dual Encoder-based \revisedcom{model} \cite{Murray_Perera_2020-trajectory-dual-encoder}. However, two key issues persist: \textbf{(1)} predictive accuracy often exhibits significant offsets (on the order of hundred meters for ships), necessitating more semantic-level predictions for decision-making; and \textbf{(2)} data and predictions primarily adopt a \revisedcom{global perspective}, lacking an ego-centric perspective\revisedcom{, which is} critical for ASV's \revisedcom{on-board} collision avoidance decision-making.

Other common approaches are the motion- and goal state-focused \revisedcom{intent} inference, primarily focusing on \textbf{COLREGs compliance} \cite{Wang_2020-colreg-inference, KAIST-intent-avoidance2022, García-Maza_Argüelles-colreg-2022, Leavitt-MIT-thesis, KAIST_RAL_2022}. The COLREGs compliance introduces some inherent challenges for those methods \revisedcom{due to:} %
\textbf{(1)} \textit{rule ambiguity}: intention inference on collision avoidance logic (\textit{give-way} or \textit{stand-on}) as a binary value was predicted and updated based on a pairwise relationship between vehicles \cite{Wang_2020-colreg-inference}. \revisedcom{A recent work introduced a} Dynamic Bayesian Network to calculate the probability of rule-compliance in the velocity space~\cite{KAIST-intent-avoidance2022}. \revisedcom{However, these approaches utilize an unclear classification of} encounter situations (e.g., a geometric boundary for head-on vs. crossing is not explicit \revisedcom{as denoted in} \cite{García-Maza_Argüelles-colreg-2022}) due to the inherent ambiguity of the rules, such that the intention inference can vary depending on the interpretation.
\textbf{(2)} \textit{reciprocal and homogeneous assumption}: intention information was used and allowed an ego-vehicle to relax COLREGs \cite{Leavitt-MIT-thesis}, but \revisedcom{all} obstacles \revisedcom{were} assumed to follow homogeneous behaviors and \revisedcom{did} not have state uncertainty. \textcite{KAIST_RAL_2022} used a \textit{reciprocal} evasive algorithm proportional to the inference of the rule compliance by obstacles. The previous work with homogeneous setup raises the need for \revisedcom{algorithms that can handle heterogeneity}: homogeneous behaviors, \revisedcom{which are rarely observed in reality}, could potentially fail to meet \textit{proactive} requirements \revisedcom{of the rule}, because the ego ASV waits for others' compliance.

\invis{
\begin{enumerate}

\end{enumerate}
}

Our study's primary insight is to focus on high-level (\textit{topological}) passing intention for active intention-aware obstacle avoidance, distinguishing our approach from previous efforts, including ours \cite{jeong-2020-iros, Jeong-MOA-2022}. This strategy is enhanced by a real-world data-driven, learning-based prediction model. The marine domain's unique characteristics and its rules of the road prompt us to question, ``How will other vehicles pass with respect to my vehicle, and how can I safely navigate past them by my action?" This ego-centric and topological perspective differs from conventional trajectory-level predictions, i.e., sequences of geometric points. Moreover, we do not assume that an ego-vehicle and other obstacles utilize a reciprocal algorithm for evasion. To create a more realistic scenario, we consider obstacles that exhibit either cooperative or non-cooperative behaviors, which may differ from the ego vehicle's behavior. In previous research \cite{jeong-2020-iros}, we explored efficient local avoidance from a relative ego-centric viewpoint for a ship domain, while addressing multiple obstacles sequentially. On the other hand, in \cite{Jeong-MOA-2022}, we proposed holistic multiple obstacle avoidance, though without considering the passing intentions of other ships and collision avoidance decision-making, accordingly. In this study, our \textit{proactive} actions prioritize safety in line with the primary principles of maritime navigation and follow its semantic and topological interpretations of collision avoidance.

\invis{
\section{Problem Formulation} \label{sec:prob}
\todo{rephrase, formal form, combine with above para}
The goal of the controlled ASV is (1) to reduce the uncertainty of the encounter situation by making an informative observation of nearby an obstacle; and (2) to ensure a collision avoidance with respect to the obstacle with multi-objectives including the safety, efficiency, and information gain. We formulate this goal as maximizing the total mutual information between the prior knowledge and the expected knowledge in consideration of the measurements by the sensors to detect obstacles, such as RADAR, LIDAR.  

... similar design ... is smaller than a predefined ratio of the entropy of the initial belief prior --  so that we can also use it for reasonable threshold
}

\begin{figure}[t!]
    \centering
    \begin{minipage}[t]{0.45\columnwidth}
        \begin{subfigure}[b]{\textwidth}            
            \includegraphics[height=1in]{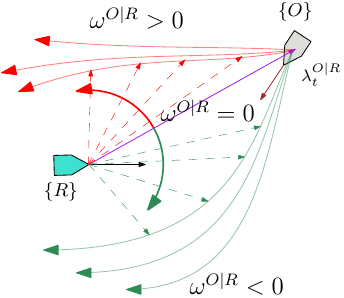}
            \vspace{-0.4em}
            \caption{}
            \label{fig:winding-number-intro}
        \end{subfigure}
    \end{minipage}
    \begin{minipage}[t]{0.45\columnwidth}
        \begin{subfigure}[b]{\textwidth}
            \includegraphics[height=1in]{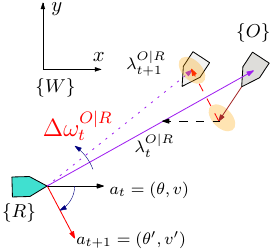}
            \vspace{-0.4em}
            \caption{}
            \label{fig:winding-number-change}
        \end{subfigure}
    \end{minipage}
    \vspace{-0.4cm}
    \caption{
    Topological classification of passing and the concept of winding number changes. \textbf{(a)} Relative view of an example scenario where $O$ approaching from the left bow of $R$ can pass on the left ($\omega^{O|R} > 0$) \revisedcom{with respect to $R$ (red lines)} or right ($\omega^{O|R} < 0$) \revisedcom{with respect to $R$ (green lines)}; and
    \textbf{(b)} action $a_t$ to $a_{t+1}$ ($\theta=\SI{90}{\degree}$ to $\theta^\prime=\SI{135}{\degree}$) makes $O$ pass as $\mathcal{P}_l$ by $\Delta \omega^{O|R} > 0$ with a fixed $v^R$ assumption. The state of each ship is $\mathbf{x}^R=[-30, 0], \theta^R=\SI{90}{\degree}, v^R=\SI{2.5}{\meter/\second}$, $\mathbf{x}^O=[0, 20], \theta^O=\SI{225}{\degree}, v^O=\SI{3.0}{\meter/\second}$, while the direction follows the maritime convention, i.e., clockwise from north.
    }
    \vspace{-0.7cm}
    \label{fig:winding-number}
\end{figure}

\section{Proposed Approach}\label{sec:approach}
The proposed approach evaluates desirable actions (heading, speed) to avoid obstacles in congested traffic while obtaining information \revisedcom{gain} \revisedcom{to actively reduce uncertainty} about the passing intent\revisedcom{ion} with respect to the ego ASV. We assume that the ego ASV has obstacle tracking information \revisedcom{via a radio frequency message reception} within a sensible range $\mathcal{S} \in \mathbb{R}^2$, like previous studies based on the Automatic Identification System (AIS) \cite{KAIST-intent-avoidance2022, Jeong-MOA-2022}, tracking for ships. \revisedcom{In the experiments, we introduce potential noise in the type-A AIS at $\SI{1}{\Hz}$, specifically adding delay in obstacle detection within $\mathcal{S}$ due to processing time for incoming obstacle data, mirroring real-world conditions \cite{solas}.} Unlike the literature, we \revisedcom{also} introduce heterogeneous uncertainty levels depending upon obstacles, commonly observed in real-world ships \cite{Emmens_2021-ais-noise}, which is tested in the experimental section. 

\subsection{Topological Classification of Passing} \label{sec:topo-class}
We \revisedcom{propose} a topological classification of passing \revisedcom{motivated by} \textbf{winding number} \cite{Kuderer_Sprunk_Kretzschmar_Burgard_2014-winding, Mavrogiannis_Balasubramanian_Poddar_Gandra_Srinivasa_2022-winding, topo-invariant-2001}. With respect to the ego ASV $R$, the progress of the obstacle $O$ \revisedcom{passing} can be categorized into the following two \textit{semantic} classes: (1) passing on the left side of the ego ASV -- $\mathcal{P}_{l}$; and (2) passing on the right side of the ego ASV -- $\mathcal{P}_{r}$. As shown \fig{fig:winding-number-intro}, the proposed concept \revisedcom{generalizes} the passing conditions, \revisedcom{illustrated as} the \textit{topologically equivalent} class of passing, while the directions and trajectories of the encounters vary. Specifically, $\mathcal{P}_{l}$ makes the ego ASV $R$ observe the obstacle $O$ with $(+)$ sign directional progress of topologically equivalent passing (counter-clockwise), whereas $\mathcal{P}_{r}$ makes $R$ observe $O$ with $(-)$ sign directional progress of passing (clockwise).

More formally, by defining $\lambda^{O|R} =\mathbf{x}^{O} - \mathbf{x}^{R}$ as the line of sight (LOS) vector for the pose of an obstacle and ego ASV $\mathbf{x}^{O}, \mathbf{x}^{R} \in \mathbb{R}^2$ in the global frame $\{W\}$, the winding number $\omega^{O|R}$ in a discrete format is:
\begin{align}
    \omega^{O|R} &= \eta \sum^T_{t=0}\Delta \omega_{t}^{O|R} = \eta \sum^T_{t=0} \Delta\theta(\lambda^{O|R}_t) \\ \nonumber
    &=  \eta \sum^T_{t=0} \atantwo(
    (\lambda^{O|R}_{t} \cross \lambda^{O|R}_{t+1}), 
    (\lambda^{O|R}_{t} \cdot \lambda^{O|R}_{t+1})
    )
\end{align}
where $\Delta \omega_{t}^{O|R} = \Delta\theta(\lambda^{O|R}_t)$ represents the change in the LOS vector angle $t$ to $t+1$, $T$ \revisedcom{is the} clearance time when $O$ safely clears away from $R$, which typically is the time to closest point of approach (TCPA) \cite{Review-NTNU2-2021} or out of sensor range, and $\eta$ \revisedcom{is a} normalization factor. The introduction of clearance time $T$ enables the sign of passing as a \textit{topological invariant}, regardless of \revisedcom{small perturbations} (e.g., zig-zag motion) of the trajectory over a short period.

Note that when $O$ approaches, if there is no appreciable change of the relative bearing, i.e., winding angle remaining $0$ (\fig{fig:winding-number-intro}), the encounter will progress to a collision (\revisedcom{consistent with the definition of \textit{collision} in} Rule 7 of COLREGs \cite{colreg}). In other words, to avoid the collision, the ASV should take an evasive action that will significantly change the LOS vector in $\{W\}$ across the time horizon (\fig{fig:winding-number-change}) and \revisedcom{our active collision avoidance approach takes such an evasive action}. Our proposed approach \revisedcom{that topologically classifies obstacle passing} is tailored to aquatic navigation such that ASVs \revisedcom{(1)} utilize a novel cost design (\sect{sec:info-gain}) and \revisedcom{active avoidance} based on \revisedcom{expected} information gain under noisy observations (\sect{sec:optimization}); \revisedcom{(2) can} choose a rule-compliant preferred action not fixed by the current passing state (e.g., $\mathcal{P}_{r}$ to $\mathcal{P}_{l}$ according to \cite{colreg}) (\sect{sec:case-study}); and (3) \revisedcom{generally consider diverse scenarios} including a vehicle overtaking the ego ASV, i.e., coming from the aft, which is different from the literature (\sect{sec:monte-carlo}).

\invis{These fit-for-purpose components differ from a previous study based on winding numbers, e.g., \cite{Mavrogiannis_Balasubramanian_Poddar_Gandra_Srinivasa_2022-winding} which assumes perfect information, vehicles only from the front, and a fixed passing state, once $O$ progresses either $\mathcal{P}_{r}$ or $\mathcal{P}_{l}$.}

\invis{
\begin{itemize}
\end{itemize}
}

\subsection{Intention Awareness Neural Network} \label{sec:lstm}
\begin{figure}[t]
    \centering
    \includegraphics[width=.85\columnwidth]{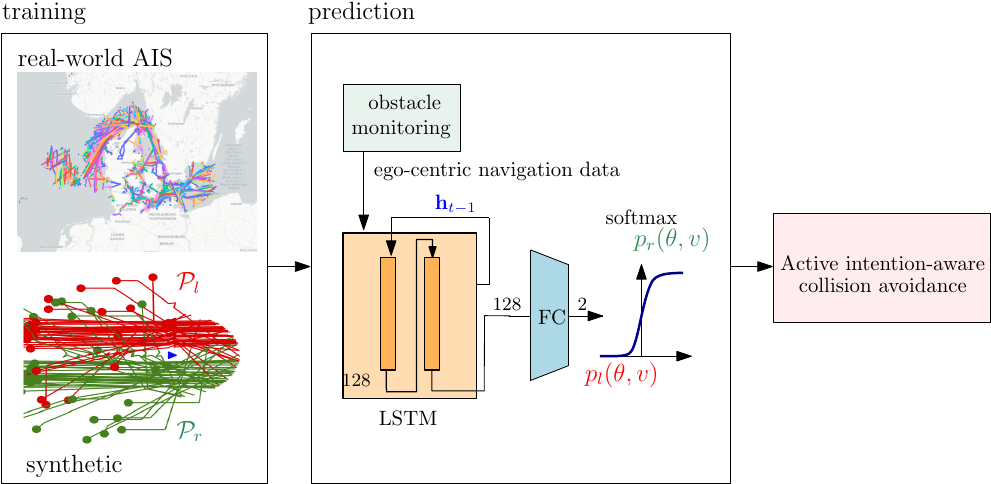}
    \vspace{-0.2cm}
    \caption{\revisedcom{LSTM-backbone neural network structure for passing classifier.}}
    \label{fig:system-architecture}
    \vspace{-0.7cm}
\end{figure}

We create the intention-awareness neural network architecture with an \textbf{LSTM} \cite{HochSchm97} backbone to perform time-series feature extraction and a fully-connected layer to classify the obstacle passing as $\mathcal{P}_l$ or $\mathcal{P}_r$ with respect to the ego ASV (\fig{fig:system-architecture}). LSTMs can accept variable length inputs in the time domain, making them optimal for our use case that requires relevant feature extraction from AIS messages of an unknown number of obstacle with an unknown AIS message frequency.

\revisedcom{We designed the network consisting of 2 stacked LSTM layers with an input size of 7 features of time-series obstacle dynamic data from ego-centric view -- i.e., $\{\mathbf{x}^{O}-\mathbf{x}^{R}, ||\mathbf{x}^{O} -\mathbf{x}^{R}||, \sin(\psi^O), \cos(\psi^O), \sin(\lambda^{O|R}), \cos(\lambda^{O|R}))\}$, where $\mathbf{x}^{O} - \mathbf{x}^{R} \in \mathbb{R}^2$ -- 128 features in the hidden state, and a fully-connected layer classifier that predicts the probability of passing $\mathcal{P}_l$ and $\mathcal{P}_r$ after the softmax activation.} \revisedcom{Our network architecture was designed to balance the trade-off between deploying an edge-capable network that can operate in real time on a robot and learning a function of sufficient complexity to excel on the dataset. This design ensures that the architecture can run on a robot's CPU in real-time (\SI{<100}{\milli\second}), aligning with the typical sensor frequencies onboard, such as marine RADAR (\SI{1}{\Hz}), thereby facilitating efficient robot deployments.}

To prepare the training data, we used \textit{real-world historical maritime AIS} and augmented it with \revisedcom{\textit{synthetic data} to balance and supplement diverse encounters (e.g., right to right passing) by \revisedcom{the} randomization scheme as shown in \fig{fig:system-architecture}}. \invis{\revisedcom{Furthermore, in the synthetic data, we include ASV-scale data, re-balance the passing cases (the majority of real-world cases are right-passing to conform with maritime rule), and simulated risky scenarios that would be unsafe to demonstrate in the real world.}to improve the robustness of the classifier to these various scenarios.} AIS data contain\revisedcom{s} (1) the vessel's static state, such as id, name, type, and dimensions; and (2) the dynamic state, such as position, heading, and speed. \revisedcom{Using} open source AIS~\cite{danish-ais}, we pre-processed one-month data (over \SI{100}{GB}) randomly chosen between $2017-2023$. We extracted vessels only with \revisedcom{the} `underway using engine' navigation status. \revisedcom{Due to the sparsity and irregular time intervals of AIS data \cite{ais-pre-process-2018}, we resampled and interpolate\revisedcom{d} the dynamic information at \SI{1}{\Hz}.} Then, if two vessels approached each other within a distance and time \revisedcom{-- TCPA (Time to Closest Point of Approach) $\leq10$ minutes; distance at TCPA $\leq3$ nautical miles --} we consider that a valid encounter (\cite{KAIST_RAL_2022, Jeong-MOA-2022}). We label the passing classification \revisedcom{($\mathcal{P}_l$ or $\mathcal{P}_r$)} from the ego-centric view as introduced in \sect{sec:topo-class}. The dataset consists of a training set (\num[group-minimum-digits={3},group-separator={,}]{37345} targets), a validation set (\num[group-minimum-digits={3},group-separator={,}]{10900} targets), and a test set (\num[group-minimum-digits={3},group-separator={,}]{4882} targets).

Then, we train\revisedcom{ed} the model based on \revisedcom{the ego-centric features extracted from} time-series AIS data \revisedcom{using binary cross-entropy loss, the Adam optimizer, and a step learning rate scheduler.}\invis{We use\revisedcom{d} time-series features based on ego-centric \revisedcom{dynamic properties} (\revisedcom{position, heading}) of the ego and obstacle vehicle to extract \revisedcom{ego-centric} information (relative position, relative bearing, relative distance).} In the inference case, the intention awareness architecture was deployed with a fixed time horizon of AIS messages received from obstacles. We chose the horizon as $\SI{10}{\second}$ for the expiration of AIS messages based on the ASV size in the experiment section, to provide an appropriate balance between the recent and past trajectory information; \revisedcom{however, users can change as per their vehicle characteristics.} \revisedcom{We are \textbf{open-sourcing} all the labeled data, pre-processing, training, evaluation code, and model weights where more details such as specific hyperparameters can be found.}\footnote{\url{https://github.com/dartmouthrobotics/passing_intention_lstm}\label{passing_intention_lstm} }

\invis{Given the small size of the LSTM model for edge deployment on a CPU, the model could also have been trained without specialized hardware.} 

\invis{
The real-world historical data serve as training for realistic scenarios, while the synthetic data augment data balance and supplement diverse encounters by randomization scheme.
For training of the passing intention inference, we used historical data from maritime AIS as well as synthetic data (\todo{See Fig x}). The real-world historical data serve as training for realistic scenarios, while the synthetic data augment data balance and supplement diverse encounters by randomization scheme. The AIS data contain (1) the static state of the vessel, such as the vessel id, name, type, and dimensions; and (2) the dynamic state, such as position, heading, and speed. Based on open source AIS\footnotemark, we pre-processed one-month data (over \SI{100}{GB}) randomly chosen between $2017-2023$. We extracted vessels only with `underway using engine' as a valid navigation status. Then, if two vessels approached each other within a distance and time (\todo{explain} TCPA, DCPA \cite{KAIST_RAL_2022, Jeong-MOA-2022}), we consider that there is a valid encounter situation and label the passing classification from the ego-centric view. The passing classification follows the topological class ($\mathcal{P}_{l}$, $\mathcal{P}_{r}$) defined in \sect{sec:topo-class}. in response to rare extreme scenarios, e.g., more extreme maneuver}

\invis{
\begin{itemize}
\end{itemize}
}

\subsection{Information Gain-driven Action Evaluation} \label{sec:info-gain}
Based on the topological class of passing introduced in \sect{sec:topo-class}, we propose \revisedcom{a novel} action cost that covers \revisedcom{the} determination of passing \textit{direction} and \textit{magnitude} \revisedcom{in our} active strategy of the ego ASV to reduce the uncertainty of the passing intention of an obstacle. From $t$ to $t+1$, we propose an approach based on ``How fast can the ego ASV change the winding angle of an obstacle in order to avoid collision?" -- see \fig{fig:winding-number-change}. Intuitively, when the obstacle's passing intention is uncertain, a proactive action by the ego ASV that changes larger winding angle $|\Delta \omega^{O|R}|$ within a fixed time is more effective in reducing uncertainty, such that the \textit{relative bearing} can progress to $(+)$ or $(-)$ \revisedcom{more rapidly}, i.e., without a collision; \revisedcom{otherwise,} the winding angle remains $0$ \revisedcom{in case of a collision}.

\subsubsection{Single encounter} 
\begin{figure}[t!]
    \centering
    \begin{minipage}[t]{0.48\columnwidth}
        \begin{subfigure}[b]{\textwidth}
            \includegraphics[width=\linewidth]{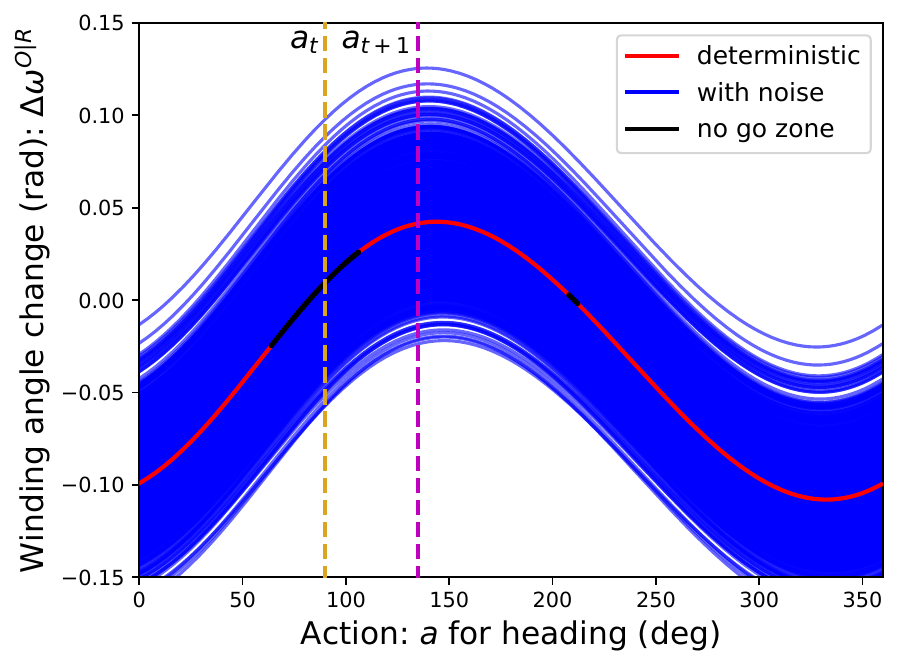}
            \vspace{-0.6cm}
            \caption{}
            \label{fig:winding-sign}
        \end{subfigure}
    \end{minipage}
    \begin{minipage}[t]{0.41\columnwidth}
        \begin{subfigure}[b]{\textwidth}
            \includegraphics[width=\linewidth]{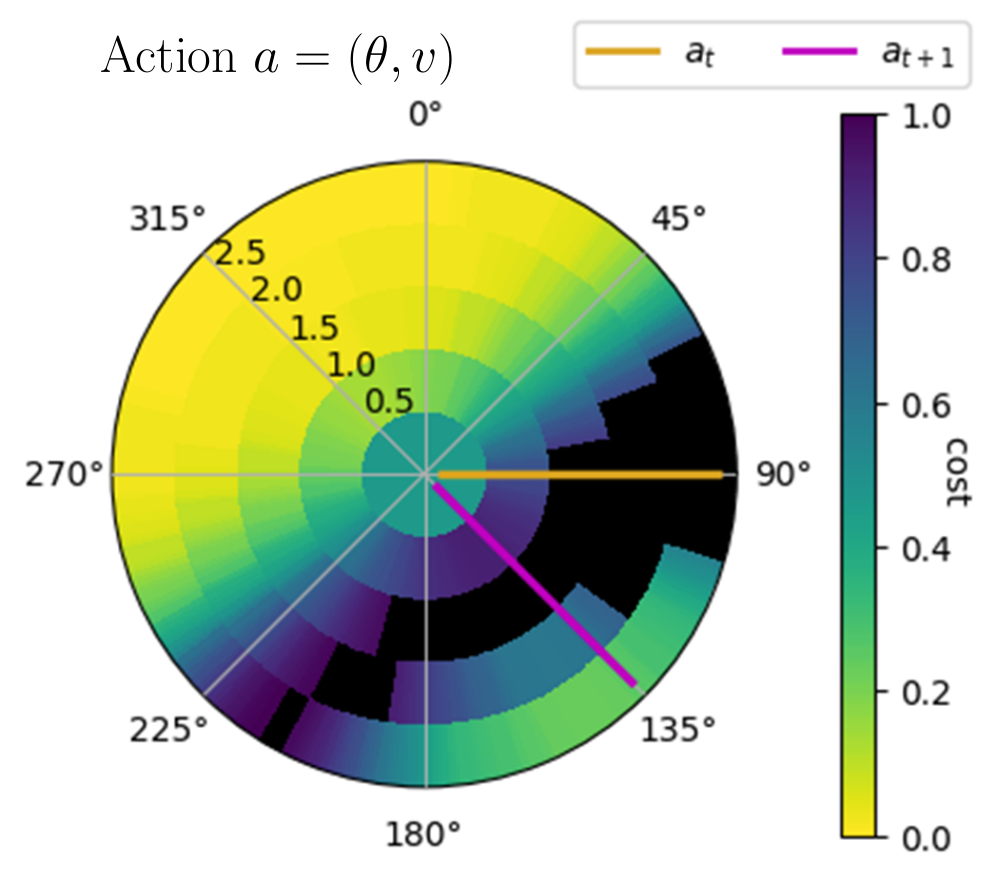}
            \vspace{-0.6cm}
            \caption{}
            \label{fig:info-gain}
        \end{subfigure}
    \end{minipage}
    \vspace{-0.4cm}
    \caption{Winding number changes and \revisedcom{expected} information gain \revisedcom{based on a next action in \fig{fig:winding-number} scenario}. \textbf{(a)} $\Delta \omega^{O|R}$ under a deterministic and noisy condition with sampling $M=1,000$ with noise in \sect{sec:result}. $a_{t+1}$ has more distribution of $\Delta \omega^{O|R} > 0$, i.e., higher $p_l$ than $a_{t}$ \revisedcom{with a fixed $v$ assumption}; and \textbf{(b)} $\Tilde{I}$ cost shows $a_{t+1}$ with better information gain (less uncertainty) than $a_t$ while $a_t$ belongs to the \textit{no-go-zone} (black). Note that the best \revisedcom{information gain} occurs at $\SI{315}{\degree}$ while it directs to the opposite direction to the current destination, which is handled by multi-objective optimization in this study.
    }
    \vspace{-0.7cm}
    \label{fig:winding-number-graph}
\end{figure}

In a deterministic scenario, the changes in winding angle are depicted in Figure \ref{fig:winding-sign} (red \revisedcom{line}). For a realistic scenario, we introduce noise in the pose $(x, y)$, speed, and heading of the obstacle $O_i$, which follows a zero-mean Gaussian distribution with standard deviations $\sigma^i_{x}$, $\sigma^i_{y}$, $\sigma^i_{\theta}$, $\sigma^i_{v}$, as observed by AIS. We then \revisedcom{sample} $M$ number of particles for the measurement. Expected changes in winding angle by \revisedcom{the next} action of the ego ASV are shown in Figure \ref{fig:winding-sign} (blue \revisedcom{lines}). We map the probabilities of $O_i$ passing to the left \revisedcom{and} right as $p^i_l$ \revisedcom{and} $p^i_r$, respectively, \revisedcom{as per} the ego ASV's action $a = (\theta, v)$ using $M$ particles, where $\theta$ and $v$ represent the ego ASV's heading and speed. For a feasible action $a_{t+1}$ at the next time step $t+1$, we define the \textbf{entropy} of \revisedcom{an obstacle} $O_i$ passing using Shannon entropy \cite{Shannon-entropy-1948} as:
\begin{equation} \label{eq:expected-entropy}
    H(\mathcal{P}^i_{t+1} | a_{t+1}) = - \sum p^i_{\textit{dir}}(a_{t+1}) * \log\, p^i_{\textit{dir}}(a_{t+1}) 
\end{equation}
where $p^i_{\textit{dir}}$ is the probability of $O_i$'s passing in a certain direction ${\textit{dir}}\in\{l,r\}$. Based on the past history of the ego vehicle's actions $a$ and observations $z^i$ for $O_i$, we also define the probabilities of passing $p^i_l(a_{t_h:t})$ and $p^i_r(a_{t_h:t})$ using the LSTM, along with the corresponding entropy:
\revisedcom{\begin{equation} \label{eq:current-entropy}
    H(\mathcal{P}^i_t | z^i_{t_h:t},a_{t_h:t}) = - \sum p^i_{\textit{dir}}(a_{t_h:t}) * \log\, p^i_{\textit{dir}}(a_{t_h:t}) 
\end{equation}
where $t_h$ represents the sliding window for monitoring. Note that $H(\mathcal{P}^i_t | z^i_{t_h:t},a_{t_h:t})$ in \eq{eq:current-entropy} represents the \textit{current entropy} based on information up to time $t$, whereas $H(\mathcal{P}^i_{t+1} | a_{t+1})$ in \eq{eq:expected-entropy} serves as the \textit{expected entropy} at time $t+1$.} Without the LSTM-backbone architecture, the probabilities $p^i_l(a_{t_h:t})$ and $p^i_r(a_{t_h:t})$ can be $p^i_l(a_{t})$ and $p^i_r(a_{t})$ by looking at only the information \revisedcom{at the current timestamp $t$}. We perform an ablation study to observe the impact of removing the LSTM\revisedcom{-based past history analysis}.

Finally, we define \textbf{information gain} to reduce the uncertainty of obstacle passing as follows:
\begin{equation} \label{eq:info-gain}
    I(\mathcal{P}^i_{t+1}) = H(\mathcal{P}^i_t | z^i_{t_h:t},a_{t_h:t}) - H(\mathcal{P}^i_{t+1}| a_{t+1})
\end{equation}
Intuitively, maximizing $I(\mathcal{P}^i_{t+1})$ can return an action $a_{t+1}$ as $(\theta_{t+1}, v_{t+1})$ which the ego ASV will take, such that the passing side of the obstacle becomes more evident. To make \eq{eq:info-gain} consistent with the cost design for \textit{minimization} in \sect{sec:optimization}, we remapped $I(\mathcal{P}^i_{t+1})$ to $\Tilde{I}(\mathcal{P}^i_{t+1}) = (1 - I(\mathcal{P}^i_{t+1})) / 2$. To reduce the passing uncertainty of an obstacle while guaranteeing collision avoidance, the proposed approach evaluates the `next-best action' that achieves `active intention-aware\revisedcom{ness}' from a set of feasible actions -- see \fig{fig:info-gain}.
\invis{
\begin{equation}
    \Tilde{I}(\mathcal{P}^i) = (1 - I(\mathcal{P}^i)) / 2
\end{equation}
since both $H(\mathcal{P}^i | z_{t_h:t},a_{t_h:t})$ and $H(\mathcal{P}^i | a_{t+1})$ are bounded by $[0,1]$ for binary classification and $I(\mathcal{P}^i)$ by $[-1,1]$.}

\invis{
}

\subsubsection{Multiple encounters} To extend the information gain of obstacle passing to multi-encounter scenarios, we \revisedcom{extend} the \textit{obstacle cluster\revisedcom{ing}} proposed in our previous study \cite{Jeong-MOA-2022}. A cluster is defined as a group of static and dynamic obstacles that have similar motion attributes with respect to an ego ASV --  temporal (time to CPA; \textbf{TCPA}), spatial (distance at CPA; \textbf{DCPA}),  and angular (\textbf{relative bearing}) similarity -- such that the ego ASV should not enter an obstacle's domain as well as narrow areas between obstacles.\invis{-- \revisedcom{please} refer to \cite{Jeong-MOA-2022} for \revisedcom{the} details.}
With multiple obstacles in a cluster, the proposed algorithm calculates the information gain as follows:
\begin{equation} \label{eq:info-gain-aggregate}
    \Tilde{I}(\mathcal{P}^{C_k}_{t+1}) = \sum_{O_i \in C_k} \alpha_{i} *  \Tilde{I}(\mathcal{P}^{O_i}_{t+1})
\end{equation}
where $O_i$ is a member obstacle in a cluster $C_k$ and $\alpha_i$ is a weight coefficient for $O_i$. For each obstacle $O_i$, $\alpha_i$ is:
\begin{equation} \label{eq:info-gain-trace}
    \alpha_i = \frac
        {
            \trace(\mathbf{cov}{(
            \mathbf{w}^i_{x},
            \mathbf{w}^i_{y}, 
            \mathbf{w}^i_{\theta}},
            \mathbf{w}^i_{v}
            ))}
        {
            \trace_{\max}(\mathbf{cov}{(
            \mathbf{w}^j_{x},
            \mathbf{w}^j_{y}, 
            \mathbf{w}^j_{\theta}},
            \mathbf{w}^j_{v}
            ))
        }
\end{equation}
where $\mathbf{w}^i \sim \mathcal{N}(0,\,\sigma^i)$ is a noise vector with the standard deviation of pose $x,y$, heading, and speed of $O_i \in C_k$ represented by $\sigma^i_{x}, \sigma^i_{y}, \sigma^i_{\theta}, \sigma^i_{v}$, $\mathbf{cov}(\cdot)$ \revisedcom{is} a covariance matrix, $\trace(\cdot)$ \revisedcom{is} a trace of a matrix \revisedcom{(the sum of the square of variances)}, and $\trace_{\max}$ \revisedcom{is} a maximum trace among traces of member obstacle $O_j \in C_k$. \revisedcom{$O_j$ is the obstacle with the highest uncertainty, where $j\neq i$.} Intuitively,\invis{the proposed cost design is aligned with the minimization $\Tilde{I}(\mathcal{P})$ so that} an obstacle with greater uncertainty has a higher cost than another obstacle with less uncertainty.

\invis{
where $\mathbf{w}_{x} \sim \mathcal{N}(0,\,\sigma^i_{x})$, $\mathbf{w}_{y} \sim \mathcal{N}(0,\,\sigma^i_{y})$, $\mathbf{w}_{\theta} \sim \mathcal{N}(0,\,\sigma^i_{\theta})$, $\mathbf{w}_{v} \sim \mathcal{N}(0,\,\sigma^i_{v})$  represent a noise vector for speed and heading, $\mathbf{cov}(\mathbf{w}_{x}, \mathbf{w}_{y},\mathbf{w}_{v}, \mathbf{w}_{\theta})$ a covariance matrix by $\mathbf{w}_{x}, \mathbf{w}_{y}, \mathbf{w}_{v}, \mathbf{w}_{\theta}$ vector, $\trace(\cdot)$ a trace of a matrix, and $\trace_{max}$ a maximum trace among traces of member obstacle $O_j \in C_k$.}

Finally, with multiple \textit{clusters}, the  aggregated information gain from individual clusters is derived by extending \eq{eq:info-gain-aggregate}: $\Tilde{I}(\mathcal{P}_{t+1}) = \sum_{C_k} \beta_{k} * \Tilde{I}(\mathcal{P}^{C_k}_{t+1})$ where $\beta_k = \max(\mathbf{a})$ and $\mathbf{a}$ is a vector composed of $\alpha_i$ for $O_i \in C_k$.

\invis{
\begin{itemize}
\end{itemize}
}

\subsection{Multi-Objective Optimization for Active Avoidance} \label{sec:optimization}
To find the optimal heading and velocity actions $\theta^*, v^*$ for the ASV, we extend the multi-objective optimization we proposed in~\cite{Jeong-MOA-2022} to include the information gain just described, as an additional criterion (marked in {\color{OliveGreen}green}):
\begin{equation} 
\small
\label{eq:weight-sum}
    (\theta^*, v^*) = \argmin_{\theta, v \in \mathcal{A} - \mathcal{A}^\prime} 
\underbrace{J_{d}(\theta, v)}_\textrm{deviation}
                    + \underbrace{w_s \, J_{s}(\theta, v)}_\textrm{safety} 
                    +  
                    {\color{OliveGreen}\underbrace{w_i \, J_{i}(\theta, v)}_\textrm{information gain} }
\end{equation}

\invis{
We formulate a multi-objective optimization to identify the optimal action from the action space $\mathcal{A}$ of a controlled ASV. 
}
The set of possible actions $\mathcal{A}$ is a discrete grid by a combination of heading and speed, represented by $\theta$ ($[0, 360)$ with a $\SI{1}{\degree}$ step) and $v$ (ratio $[0,1]$ of the maximum target speed with a $0.25$ step), respectively. \revisedcom{Within $\mathcal{A}$, we define \textit{no-go-zone} action boundary $\mathcal{A}^\prime$, which is determined using the concept of a virtual \textit{ship domain} as described in our prior research \cite{jeong-2020-iros}. This ship domain is divided into two distinct regions: \textit{collision boundary} $\mathcal{C}$, which is an area ASVs are forbidden to enter due to it being deemed a collision, even in cases where passing without physical contact might seem possible; and \textit{risky boundary} $\mathcal{R}$, an area where ASVs may enter but must exercise increased caution to maintain safety. $\mathcal{A}^\prime$ is specifically defined by the margins of evasive actions with respect to $\mathcal{C}$ of an obstacle. \invis{In practical terms, it refers to any evasive action that extends to a line tangent to $\mathcal{C}$, viewed from the ASV's perspective. The margin can be calculated by a motion vector that leads an ASV into the edge of $\mathcal{C}$.}}

\invis{
\begin{equation} \label{eq:linearequation}
    \begin{cases}
         (\mathbf{\dot{x}}^\theta_{R} - \mathbf{\dot{x}}_{O}) \cross (\mathbf{x}_{P}^\theta - \mathbf{x}_{O}) = \mathbf{0} & \\ 

        \norm{\mathbf{\dot{x}}^{\theta}_{R}} = \norm{\mathbf{\dot{x}}^{\theta_{0}}_{R}} & \\       
    \end{cases}
\end{equation}
\noindent where $\mathbf{x}_P^\theta$ is a tangent point of $\mathcal{C}$ when the ASV's heading is $\theta$, $\theta_0$ is the current heading, and  $\mathbf{\dot{x}}^{\theta}_{R}$, $\mathbf{\dot{x}}^{\theta_{0}}_{R}$ is velocity of ASV for $\theta$, $\theta_{0}$, respectively.
}

\invis{
The cost of actions are evaluated as weighted sum of multiple objectives as follows:
\begin{flalign}
 \label{eq:weight-sum}
    & J(\theta, v) = \underbrace{J_{d}(\theta, v)}_\textrm{deviation}
                    + \underbrace{w_s \, J_{s}(\theta, v)}_\textrm{safety} 
                    +  \underbrace{w_i \, J_{i}(\theta, v)}_\textrm{information gain} , %
\end{flalign}

where $(\theta, v)$ is an action $a \in \mathcal{A} - \mathcal{A}^\prime$ to be evaluated;
}
\textbf{(a)}   $J_{d}(\theta, v) =  w_f \, f(\theta) 
                        + w_{f_{2}} \, f_2(\theta) 
                        + w_g \, g(v)$  is a deviation cost from a desired goal\invis{$a = (\theta, v)$} represented by $f, f_2, g$, respectively; $f(\theta)$ is based on $\theta_{wp}$, \revisedcom{$g(v)$ is based on $v_{wp}$ where $\theta_{wp}$ and $v_{wp}$ is the desired heading and speed to the next waypoint, respectively. $f_2(\theta)$ is based on $\theta_{tgt}$ where $\theta_{tgt}$ is a local target heading goal to preventing chattering \cite{mpc-Hagen2018, Jeong-MOA-2022}} in relation to hysteresis, while avoiding obstacles; \textbf{(b)} $J_s$ is a safety level cost based on DCPA in this study, i.e., safe distance off at the closest approach; \textbf{(c)} $J_i$ is a  cost related to the information gain about the obstacle passing intention by $\Tilde{I}(\mathcal{P}| \theta,v) $ introduced in \sect{sec:info-gain}, \revisedcom{which is the core part of this study, for \textbf{active avoidance} by intention-awareness\invis{Intuitively, $J_i$ cost represents: `how much information gain on the passing intention of obstacle(s) an evaluated action will obtain}}; and all $w$ are related weights.
\section{Results and Evaluation}\label{sec:result}

We quantitatively evaluated our approach by running  \textbf{(1)} \revisedcom{a total of \num[group-minimum-digits={3},group-separator={,}]{2400}} Monte Carlo simulations\invis{in randomized environments with various encounters and other agents' behaviors and dimensions,}. %
We also carried out \textbf{(2)} \revisedcom{real robot experiments in the Caribbean Sea thus including real-world disturbances} and \textbf{(3)} real maritime accident case study. For the ego ASV \revisedcom{during both simulation and real-world experiments}, we \revisedcom{tested with} our custom ASV \catabot. \catabot\, has \SI{2.5}{\meter} length, \SI{1.4}{\meter} beam, \SI{100}{\meter} for the sensing range, \revisedcom{with dynamic characteristics} as maximum linear and angular speed \SI{2.5}{\meter/\second}, \SI{45}{\degree/\second}, respectively. \revisedcom{We used a computer equipped with an Intel i7-7820X 8-core \SI{3.6}{GHz} processor, \SI{32}{GB} RAM, and NVIDIA GPU RTX 3090 Ti with \SI{24}{GB} VRAM.} \revisedcom{\catabot~is equipped with a NVIDIA Orin Jetson-Small Developer Kit 12-core Arm Cortex 64-bit CPU, \SI{32}{GB} RAM, and 2048-core NVIDIA Ampere architecture GPU with 64 Tensor cores.} \revisedcom{For the neural network training, the model achieved a F1 score of $0.9256$. The details about training, validating and testing can be found in our open-sourced repository\footref{passing_intention_lstm}.}

\invis{and 3D simulator \cite{gazebo_2004} with open-sourced ASV model \cite{robot-x} (about \SI{6}{\meter} length).} 
\invis{real-world-like experiments on a realistic 3D simulator including wave and wind disturbances, for a qualitative evaluation.}

\subsection{Experimental Setup} \label{sec:monte-config}
We performed Monte Carlo simulations \revisedcom{binned} by the set of obstacle numbers $\{10, 20, 30\}$ with $100$ environments per method (\num[group-minimum-digits={3},group-separator={,}]{1500} runs) \revisedcom{with additional ablation study (\num{900} runs).} \invis{and \revisedcom{an} additional ablation study for rule-compliance with (\num{900} runs).}The test area is within \qtyproduct{200 x 200}{m}, while obstacles' size, speed, and encounter directions were randomly chosen. The start and goal positions were set as $[0, -100], [0, 100]$, \revisedcom{respectively}. The baseline methods are Velocity Obstacle (VO)-based \cite{VO-COLREG-kuwata}, and Multiple Obstacle Avoidance (MOA)-based \cite{Jeong-MOA-2022}) that have \revisedcom{previously} shown state-of-the-art performance in multiple encounters. The active intention-aware approach proposed in this paper is termed MOA$^+$LSTM.
\revisedcom{As a part of the ablation study,} we term  MOA$^+$ \revisedcom{by} ablating the proposed LSTM and having a prediction based only on the current information rather than the history (\sect{sec:info-gain}). \revisedcom{Furthermore, to} observe the impact of information gain, we modify VO by considering individual obstacles in Eq. (\ref{eq:info-gain-aggregate}), (\ref{eq:info-gain-trace}), not as a group -- we call it VO$^+$.
\revisedcom{In summary, we compare our proposed approach (MOA$^+$LSTM) with $\{$MOA$^+$, MOA, VO$^+$, VO$\}$.
We tuned the essential parameters (\sect{sec:optimization}) with separate $50$ scenarios.} 

In each scenario, we selected action schemes for other vehicles, \revisedcom{inspired by \cite{Alonse-Mora_2021-where-to-go}, as follows}: (1) $80\%$ \textit{non-cooperative} and $20\%$ \textit{cooperative}; (2) fully \textit{non-cooperative}. Non-cooperative vehicles followed a constant velocity (CV) motion, exhibiting limited responsiveness frequently observed in real-world scenarios \cite{shah-2014-cv}. Cooperative vehicles, instead, executed evasive actions; however, we do not assume that such vehicles use a \textit{reciprocal} or \textit{coordinated} avoidance strategy, with respect to any other obstacle including the ego ASV. Specifically, for a cooperative agent, we randomly selected a state-of-the-art local planner, from \revisedcom{Artificial} Potential Field (APF)-based \cite{XUE-APF-2011}, Dynamic Window Approach (DWA)-based \cite{Fox_Burgard_Thrun_DWA1997}, VO-based \cite{VO-COLREG-kuwata}, and MOA-based \cite{Jeong-MOA-2022} to mimic real-world conditions consisting of \revisedcom{heterogeneous behaviors by traffic ships. Note that for each scenario across ego-vehicle's avoidance strategies $\{$MOA$^+$LSTM, MOA$^+$, MOA, VO$^+$, VO$\}$, we made a selection of cooperative obstacle's strategy consistent for fair comparison.} 

Moreover, we ran trials with and without AIS noise \revisedcom{to demonstrate robustness in both ideal and real-world-like scenarios}. AIS noise for vehicle $i$'s pose $x, y$ ($\sigma^i_x$ and $\sigma^i_y$), heading ($\sigma^i_{\theta}$), and speed ($\sigma^i_{v}$) is from uniform distributions: $U \sim (0, 0.3)\, \SI{}{\meter}$, $U \sim (0, 0.3)$ radians, and $U \sim (0, 0.5)\,\SI{}{\meter/\second}$, respectively, based on the vehicle's dimension and literature \cite{KAIST-intent-avoidance2022}. Then, \revisedcom{we add the noise to the corresponding vehicle's AIS broadcast}, with $\mathbf{w^i} \sim \mathcal{N}(0,\sigma^i)$
\revisedcom{for} $\SI{1}{\Hz}$ frequency as type-A AIS. Ego ASV samples particles $M=1{,}000$ as described in \sect{sec:info-gain}. The heterogeneous AIS noise on each vehicle provides a more realistic scenario \cite{Emmens_2021-ais-noise} rather than the same noise levels for all vehicles. \invis{To ensure a fair performance analysis, we used a fixed random seed for methods \revisedcom{per} each scenario.
}

\begin{table}[]
\centering
\caption{Comparison of overall performance of collision avoidance: success rate including nearmiss contact.}
\vspace{-0.3cm}
\label{tab:eval-overall}
\resizebox{\columnwidth}{!}{%
\begin{tabular}{c|cc|ccccc}
\hline
\multirow{2}{*}{\textbf{obstacles}} &
  \multicolumn{2}{c|}{\textbf{encounter property}} &
  \multicolumn{5}{c}{\textbf{success rate}} \\ \cline{2-8} 
 &
  \multicolumn{1}{c|}{\textbf{TE$^1$ [ea]}} &
  \textbf{AET$^2$ [ea]} &
  \multicolumn{1}{c|}{\textbf{\begin{tabular}[c]{@{}c@{}}MOA\\ +LSTM$^3$\text{*}\end{tabular}}} &
  \multicolumn{1}{c|}{\textbf{MOA+$^3$}} &
  \multicolumn{1}{c|}{\textbf{MOA$^3$}} &
  \multicolumn{1}{c|}{\textbf{VO+$^4$}} &
  \textbf{VO$^4$} \\ \hline \hline
\multicolumn{1}{c|}{10} &
  \multicolumn{1}{c|}{9.98 ± 0.13} &
  3.50 ± 1.86 &
  \multicolumn{1}{c|}{\textbf{0.99}} &
  \multicolumn{1}{c|}{0.97} &
  \multicolumn{1}{c|}{0.96} &
  \multicolumn{1}{c|}{0.77} &
  0.72 \\ \hline
\multicolumn{1}{c|}{20} &
  \multicolumn{1}{c|}{19.84 ± 1.58} &
  6.91 ± 3.28 &
  \multicolumn{1}{c|}{\textbf{0.94}} &
  \multicolumn{1}{c|}{0.91} &
  \multicolumn{1}{c|}{0.90} &
  \multicolumn{1}{c|}{0.65} &
  0.65 \\ \hline
\multicolumn{1}{c|}{30} &
  \multicolumn{1}{c|}{29.77 ± 2.18} &
  10.16 ± 4.64 &
  \multicolumn{1}{c|}{\textbf{0.96}} &
  \multicolumn{1}{c|}{0.93} &
  \multicolumn{1}{c|}{0.92} &
  \multicolumn{1}{c|}{0.71} &
  0.64 \\ \hline
\multicolumn{3}{c|}{Overall} &
  \multicolumn{1}{c|}{\textbf{0.959}} &
  \multicolumn{1}{c|}{0.936} &
  \multicolumn{1}{c|}{0.926} &
  \multicolumn{1}{c|}{0.706} &
  0.668 \\ \hline
\end{tabular}%
}
\raggedright
\scriptsize{$\text{*}$: proposed method, $^1$: Total obstacle encounters from the start to the goal position, $^2$: average encounters per timestamp, $^3$: clustering-based, $^4$: individual-based}
\vspace{-0.2cm}
\end{table}

\begin{figure}[t!]
    \centering
    \begin{minipage}[t]{0.49\columnwidth}
        \begin{subfigure}[b]{\textwidth}
           \includegraphics[width=\linewidth]{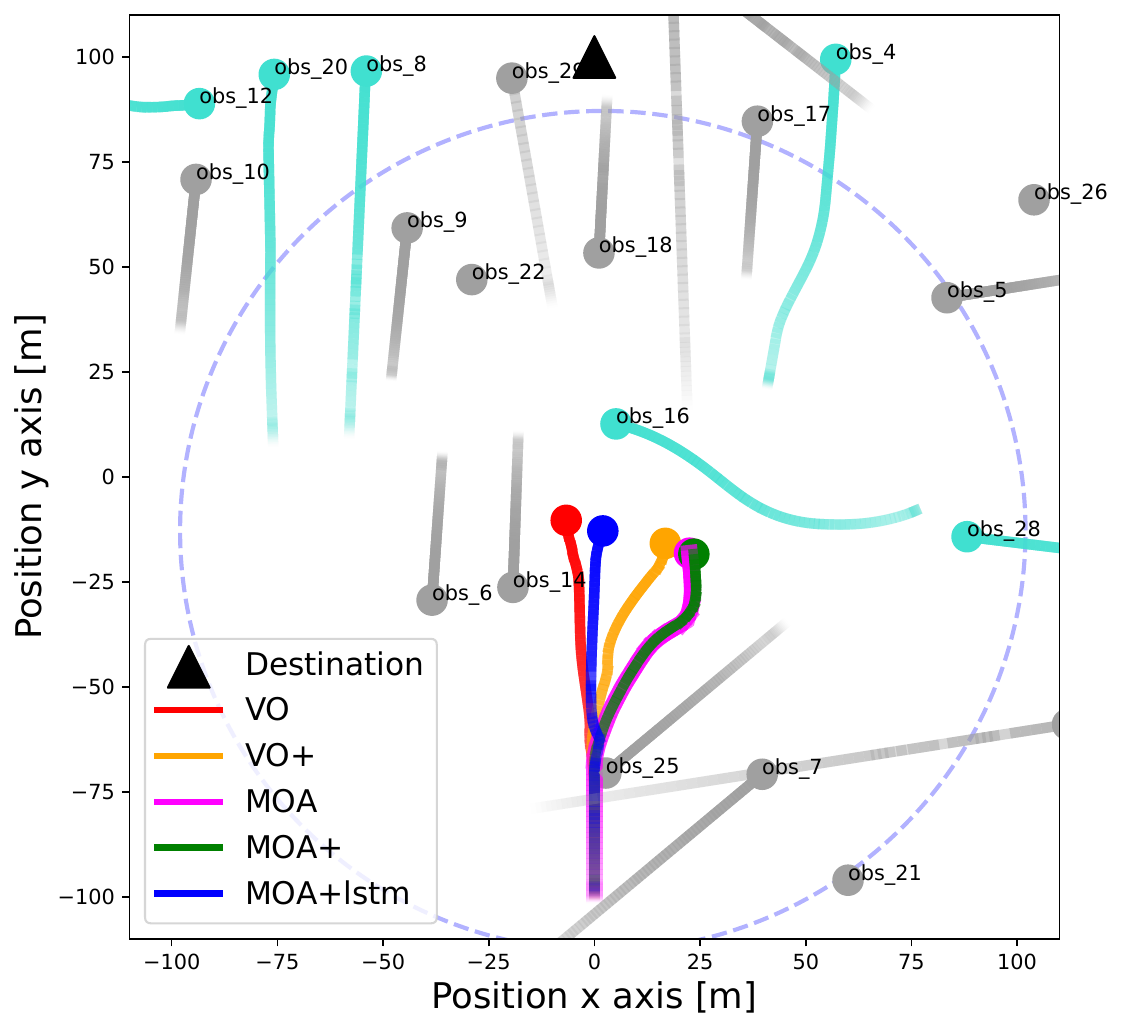}
           \vspace{-0.6cm}
           \caption{time elapsed: $ \SI{40}{\second}$}
         \end{subfigure}
    \end{minipage}
    \begin{minipage}[t]{0.49\columnwidth}
        \begin{subfigure}[b]{\textwidth}
           \includegraphics[width=\linewidth]{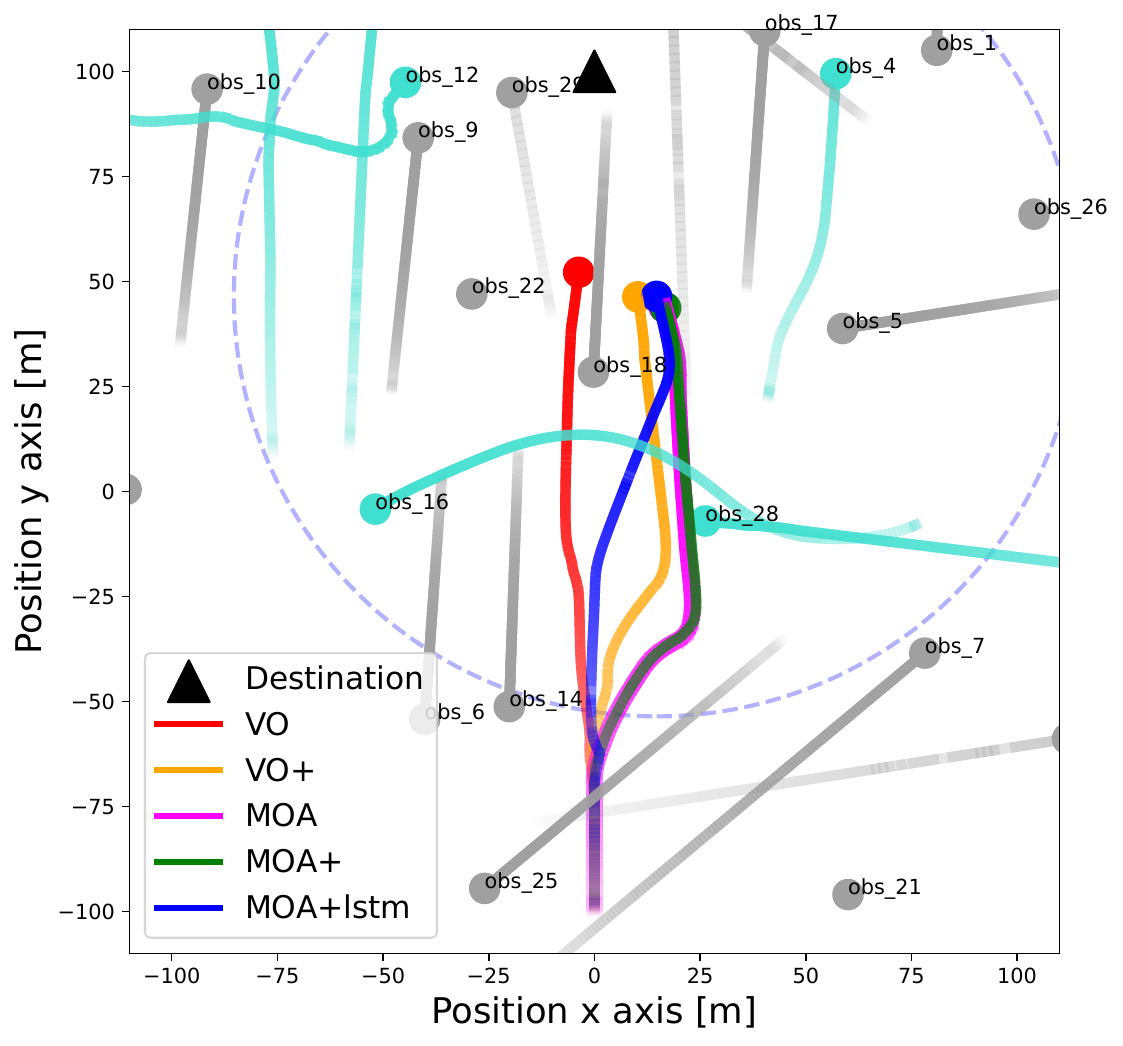}
           \vspace{-0.6cm}
           \caption{time elapsed: $ \SI{65}{\second}$}
         \end{subfigure}
    \end{minipage}
    \vspace{-0.8cm}
    \caption{\revisedcom{Comparison of trajectories in an example scenario with $30$ obstacles (cooperative obstacle: cyan, non-cooperative obstacle: gray) under ego ASV's sensible range $\mathcal{S}$ $\SI{100}{\meter}$ (only our method drawn in blue dots for clarity). Key obstacles within $\mathcal{S}$: $obs ~6, 14, 18$ head-on, $obs~5, 16, 25, 28$ crossing from the right, $obs~22$ anchored, $obs~12$ crossing from the left, $obs~9, 29$ overtaken. \textbf{(a)} After passing $obs~25$, VO$^+$, MOA, MOA$^+$ took abrupt actions with respect to $obs~16$, whereas the proposed MOA$^+$LSTM captured history of intention changes by $obs~16$ and took a smooth action to pass astern of it ($\mathcal{P}_{l}$ passing); and \textbf{(b)} After passing $obs~16$, VO entered between $obs~22$ and $obs~18$. However, the proposed MOA$^+$LSTM conducted a holistic consideration of $obs~18,22$ as a cluster, and took an active action by right-side turn to make both $obs~18,22$ as $\mathcal{P}_{l}$ passing. Note that there are other obstacles within $\mathcal{S}$ (e.g., $obs~7$), but our approach prioritizes obstacles based on ego-centric dynamic properties, as in \sect{sec:info-gain}.}}
    \vspace{-0.4cm}
    \label{fig:comparison-trajectory}
\end{figure}

\invis{
\begin{figure}[bt!]
    \centering
    \begin{minipage}[t]{0.51\columnwidth}

        \begin{overpic}[width=\textwidth]{figs/fig6-trajectory-new-new.pdf}%
            \put(20,48){\frame{\includegraphics[width=0.2\textwidth]{figs/wamv.png}}}%
        \end{overpic}
        \vspace{-0.6cm}
    
    \end{minipage}
    \begin{minipage}[t]{0.44\columnwidth}
        \begin{subfigure}[b]{\textwidth}
            \includegraphics[height=1.169in]{figs/fig6-probability.pdf}
            \vspace{-0.3cm}
            \label{fig:gazebo-probability}
        \end{subfigure}

    \end{minipage}
    \vspace{-0.2em}
    \caption{Real-world-like simulation including environmental disturbances (a) three ASVs affected by waves; (b) probability of passing predicted by LSTM with respect to ego ASV. The ego ASV passed both $O_1$, $O_2$ on the left side at about $\SI{60}{\second}$, $\SI{90}{\second}$, respectively.}
    \vspace{-2em}
    \label{fig:gazebo}
\end{figure}
}

\subsection{Performance Results} \label{sec:monte-carlo}
We evaluate the performance of the proposed MOA$^+$LSTM using quantitative metrics: (1) success rate, \revisedcom{as the most important metric}, defined as \revisedcom{the} reachability to the goal position without a physical contact and \textit{nearmiss} -- i.e., entering is not allowed, despite non-physical contact according to a definition of the ship domain \cite{Szlapczynski_Szlapczynska_2017-ship-domain, jeong-2020-iros}; and (2) computational time. \revisedcom{The overall results over a set of environments per varying number of obstacles, with/without noise, and cooperative behavior are shown in \tab{tab:eval-overall} and qualitative trajectory example in \fig{fig:comparison-trajectory}. Note that encounter property in \tab{tab:eval-overall} represents complexity of traffic.}

The proposed MOA$^+$LSTM showed the best success rate, with $0.959$ on average \revisedcom{over all runs}, while MOA$^+$, MOA, VO$^+$, and VO showed $0.936, 0.926, 0.706, 0.668$, respectively. The result also demonstrates that the proposed \textbf{active intention-aware} approach (MOA$^+$LSTM, MOA$^+$, VO$^+$) outperforms the corresponding baselines \revisedcom{without intention-aware (MOA, VO), respectively,} by including the information gain that reduces the uncertainty of passing intentions. \revisedcom{More importantly, among active intention-aware methods, the proposed MOA$^+$LSTM approach that adopts the long-term ego-centric information outperforms the current time-based approaches (MOA$^+$, VO$^+$) -- see \fig{fig:comparison-trajectory}.} \revisedcom{Note that in multiple obstacle encounters, the holistic approach that cluster\revisedcom{s} groups of nearby obstacles is found to increase the safety criteria (MOA$>$VO$^+$) aligned with our previous study \cite{Jeong-MOA-2022}.}
We found that the proposed MOA$^+$LSTM's very few nearmiss \revisedcom{cases} resulted from (1) LSTM's fixed monitoring time window, which might not fully capture a sudden course change; and (2) the approximation introduced by the sampling of the probability distribution -- \textit{finding 1}. 
Note that, if we relax the success rate considering unsuccessful navigation only those instances with a physical contact, all approaches achieved success performance over $0.97$. However, entering a ship domain -- nearmiss -- is not considered acceptable by COLREGs, \revisedcom{despite non-physical contact}.
\revisedcom{Moreover, given that \textit{risk} is a factor of \textit{frequency} and \textit{consequence}, a small percentage of improvement in the safety criteria poses a significant improvement in acceptable navigation risk for ASVs in operation \cite{risk-mass-1, risk-mass-2}.} 

Unsurprisingly, both the proposed and state-of-the-art methods showed better safety \revisedcom{performance} under \textbf{(a)} \textit{no noise} conditions than under \textit{noise} conditions. Interestingly, we note that environments with \textbf{(b)} \textit{cooperative} obstacles are not necessarily safer than environments with \textit{non-cooperative} obstacles only. The reason may be that \textit{cooperative} obstacles not only interact with our controlled ASV, but also with the rest of the obstacles, which could lead to conflicting behaviors -- \textit{finding 2}. A full in-depth analysis of safety vs.\ cooperativeness in the maritime domain is left for future work, taking inspiration from other \revisedcom{interaction-aware} research, such as \revisedcom{that for} self-driving cars \cite{Zhu_Claramunt_Brito_Alonso-Mora_2021-interaction, Li_Zhao_Xu_Wang_Chen_Dai_2021-lane-change-intention}. 

The \textit{computation time} of the proposed method \revisedcom{shows real-time performance even with a high number of obstacles ($\SI{81.4}{\milli\second} \pm \SI{35.2}{\milli\second}$ for $10$ obstacles and $\SI{124.5}{\milli\second} \pm \SI{49.0}{\milli\second}$ for $30$ obstacles). We designed our method with parallel processing of obstacle data and action evaluation modules on top of the clustering-based algorithms, which, in practice, significantly reduce the computational load.} \revisedcom{The} LSTM\revisedcom{-backbone network}'s inference time is approximately $\SI{17.2}{\milli\second} \pm \SI{14.7}{\milli\second}$. 
Overall, even with the additional computation due to the information gain and the LSTM\revisedcom{-backbone network}, our method can operate real-time, \revisedcom{ideal for on-board running on ASV with its best-in-class safety performance}.

We additionally evaluated \textit{traveled distance}, and our proposed approach is not significantly different, i.e., does not detour, compared to the state-of-the-art (ours: $\SI{212.97}{\meter}\pm\SI{12.97}{\meter}$, VO: $\SI{200.87}{\meter}\pm\SI{1.16}{\meter}$ for $30$ obstacles).

\subsection{Real-world Experiments}  \label{sec:real-robot}

\invis{
\begin{figure}
    \centering
    \begin{minipage}[t]{0.41\columnwidth}
        \begin{subfigure}[b]{\textwidth}
           \includegraphics[width=\linewidth]{figs/catabot-shore.jpg}
         \end{subfigure}
        \vspace{-0.5cm}
    \end{minipage}
    \begin{minipage}[t]{0.50\columnwidth}
        \begin{subfigure}[b]{\textwidth}
           \includegraphics[width=\linewidth]{figs/catabot-drone2.png}
         \end{subfigure}
    \end{minipage}
    \vspace{-0.3cm}
    \caption{Tested ASV model, \catabot\ (\textit{left}) and its in-water operation (\textit{right}) in the Caribbean Sea.}
    \label{fig:field-experiment}
    \vspace{-0.6cm}
\end{figure}

\begin{figure}
    \begin{minipage}[t]{0.495\columnwidth}
        \begin{subfigure}[b]{\textwidth}
           \includegraphics[width=\linewidth]{figs/config_close.pdf}
         \end{subfigure}
    \end{minipage}
    \begin{minipage}[t]{0.495\columnwidth}
        \begin{subfigure}[b]{\textwidth}
           \includegraphics[width=\linewidth]{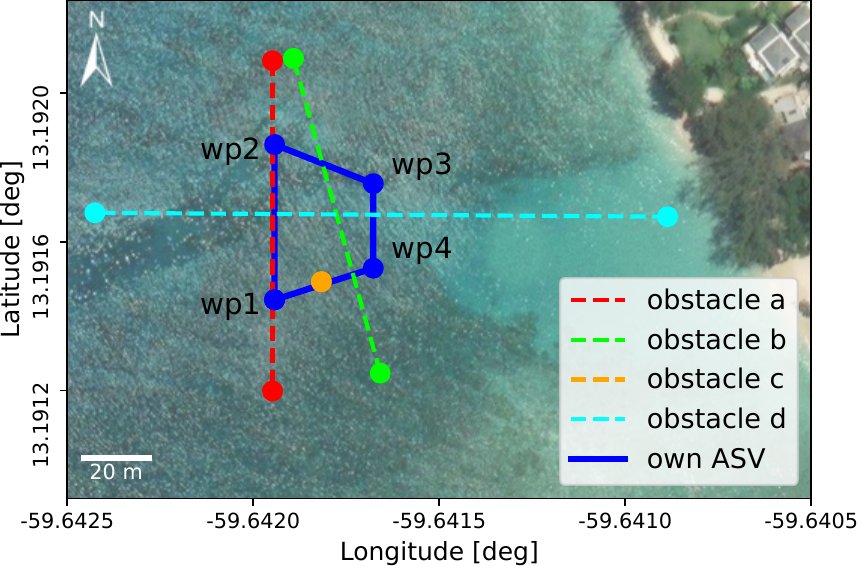}
         \end{subfigure}
    \end{minipage}
    \vspace{-0.8cm}
    \caption{Real-world experiment configuration. Location \textit{A} with \num{2} obstacles (\textit{left}) and location \textit{B} with \num{4} obstacles (\textit{right}). The distance from the shore is approximlatey $\SI{46}{\meter}$ and $\SI{108}{\meter}$, respectively.}
    \vspace{-0.3cm}
    \label{fig:field-experiment-config}
\end{figure} 
}

\begin{figure}
    \centering
    \begin{minipage}[t]{0.5\columnwidth}
        \begin{subfigure}[b]{\textwidth}
           \includegraphics[width=0.95\linewidth]{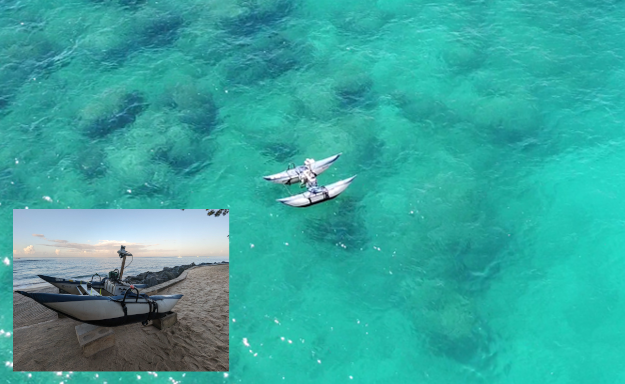}
         \end{subfigure}
    \end{minipage}
    \hfill %
    \begin{minipage}[t]{0.44\columnwidth}
        \begin{subfigure}[b]{\textwidth}
           \includegraphics[width=\linewidth]{figs/config_far.pdf}
         \end{subfigure}
    \end{minipage}
    \vspace{-0.3cm}
    \caption{Real-world experiment with our custom ASV \catabot~ in operation (\textit{left}) and experimental location \textit{B} with \num{4} obstacles (\textit{right}).}
    \vspace{-0.3cm}
    \label{fig:field-experiment-config}
\end{figure} 

\begin{figure}[t!]
    \begin{minipage}[t]{0.49\columnwidth}
        \begin{subfigure}[b]{\textwidth}
           \includegraphics[width=\linewidth]{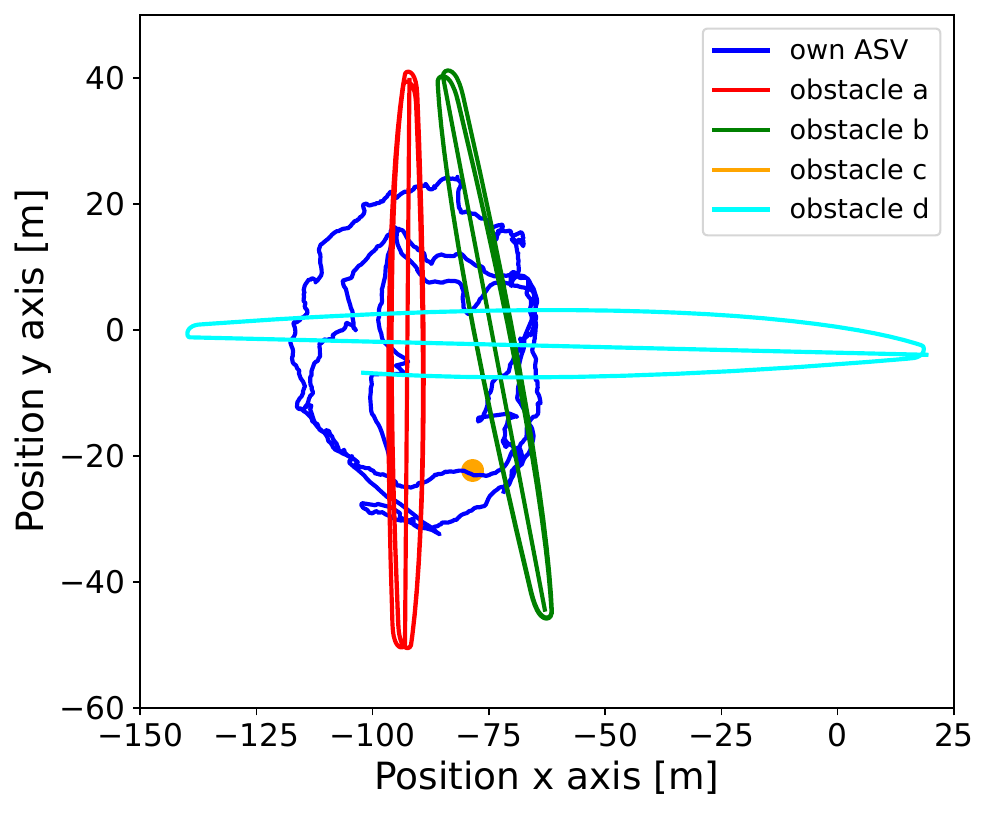}
           \vspace{-0.6cm}
           \caption{VO}
         \end{subfigure}
    \end{minipage}
    \hfill %
    \begin{minipage}[t]{0.49\columnwidth}
        \begin{subfigure}[b]{\textwidth}
           \includegraphics[width=\linewidth]{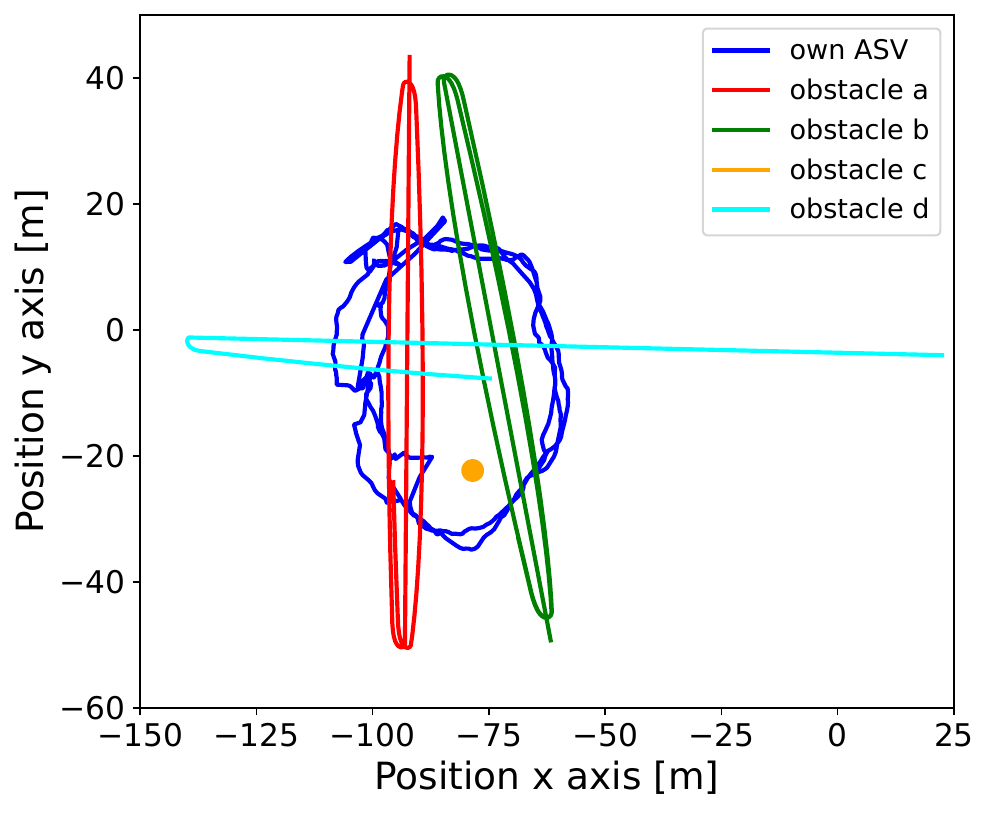}
           \vspace{-0.6cm}
           \caption{Proposed}
         \end{subfigure}
    \end{minipage}
    \invis{
    \begin{minipage}[t]{0.49\columnwidth}
        \begin{subfigure}[b]{\textwidth}
           \includegraphics[height=2.5cm,width=\linewidth]{figs/distance_VO-far.pdf}
           \vspace{-0.5cm}
           \caption{VO}
         \end{subfigure}
    \end{minipage}
    \hfill %
    \begin{minipage}[t]{0.49\columnwidth}
        \begin{subfigure}[b]{\textwidth}
           \includegraphics[height=2.5cm,width=\linewidth]{figs/distance_MOA-far.pdf}
           \vspace{-0.5cm}
           \caption{Proposed}
         \end{subfigure}
    \end{minipage}
    }
    \vspace{-0.7cm}
    \caption{Real robot experiment in location \textit{B} (far) with \num{4} obstacles. Comparison of the trajectories by the-state-of-the-art VO vs proposed MOA$^+$LSTM. (length-\si{\meter}, beam-\si{\meter}, speed-\si{\meter/\second}) -- obs. a: (2.5,1.4,0.7), obs. b: (1.5,1.0,0.5), obs. c: (1.5,0.8,0.0), obs. d: (2.5,1.4,0.5).}
    \label{fig:field-experiment-far}
    \vspace{-0.7cm}
\end{figure} 

\begin{table}[]
\centering
\caption{Performance comparison in real robot experiments.}
\vspace{-0.2cm}
\label{tab:real-perf-comparison}
\resizebox{0.7\columnwidth}{!}{%
\begin{tabular}{c|c|cccc}
\hline 
\multirow{2}{*}{\textbf{Performance}} & \multirow{2}{*}{\textbf{Method}} & \multicolumn{4}{c}{\textbf{Obstacle}}                                                    \\ \cline{3-6} 
                                      &                                  & \multicolumn{1}{c|}{a}    & \multicolumn{1}{c|}{b}    & \multicolumn{1}{c|}{c}    & d    \\ \hline \hline
\multirow{2}{*}{Nearmiss [case]} &
  Proposed &
  \multicolumn{1}{c|}{\textbf{0}} &
  \multicolumn{1}{c|}{\textbf{0}} &
  \multicolumn{1}{c|}{\textbf{0}} &
  \textbf{0} \\ \cline{2-6} 
                                      & VO                               & \multicolumn{1}{c|}{13}   & \multicolumn{1}{c|}{0}    & \multicolumn{1}{c|}{16}   & 0    \\ \hline
\multirow{2}{*}{Min. CPA [m]} &
  Proposed &
  \multicolumn{1}{c|}{\textbf{5.40}} &
  \multicolumn{1}{c|}{\textbf{9.01}} &
  \multicolumn{1}{c|}{\textbf{8.92}} &
  \textbf{16.26} \\ \cline{2-6} 
                                      & VO                               & \multicolumn{1}{c|}{2.69} & \multicolumn{1}{c|}{8.75} & \multicolumn{1}{c|}{0.63} & 6.44 \\ \hline
\end{tabular}%
}
\end{table}

We also validated the proposed approach with our custom ASV \catabot~in the real world (\fig{fig:field-experiment-config}). The experimental area is the Caribbean Sea ($\SI{13}{\degree}11'$ N, $\SI{59}{\degree}38'$ W), Barbados, with two main locations (\textit{A} and \textit{B}) on different dates. Inspired by \cite{kriso_2020}, we modeled nearby traffic \revisedcom{using} virtual obstacles because of the experimental safety.

We set up combinations of our ASV's trajectory loop (from $wp1$ to $wp4$) and the obstacle's trajectory loop (marked by the endpoints of the dotted lines) to ensure the ASV encountered a variety of traffic situations (e.g., head-on, crossing) under a repetitive scheme. \invis{This setup is particularly useful for validating the ASV's capability in real-world task scenarios like environmental monitoring. }
For a fair comparison, we applied same parameters across the methods, e.g., AIS noise, as \sect{sec:monte-config}. We compared the proposed method with the VO method across several obstacle encounters within these loops.  Due to space limitations and the similarity of outcomes, we only report results for location \textit{B}.

\invis{At location \textit{A} (about $\SI{46}{\meter}$ from the shore) with \num{2} obstacles, our ASV navigated along the loop \num{4} times under the wind direction $SE$ and speed $\SI{4.4}{\meter/\second}$.} 

At location \textit{B} %
with \num{4} obstacles, own ASV navigated along the loop \num{3} times (\fig{fig:field-experiment-far}), under wind direction $E$ and speed $\SI{8.0}{\meter/\second}$. As shown in \tab{tab:real-perf-comparison}, our proposed method with intention awareness outperformed the state-of-the-art VO method in terms of the \textit{safety} of navigation with larger CPA distance\revisedcom{s} and no nearmiss encounters. The VO-based trajectory shows more zig-zag motions under real-world noisy environments with difficulty in determining the action, while the entire mission time became longer and the safety criterion was not met (e.g., with respect to obstacle $c$). On the other hand, the proposed approach handles the noisy situation and takes \textit{proactive and safe} actions to explicitly determine the passing side of approaching obstacle(s), i.e., probability near $1.0$.

\invis{
\begin{figure}[t!]
    \begin{minipage}[t]{0.47\columnwidth}
        \begin{subfigure}[b]{\textwidth}
           \includegraphics[width=\linewidth]{figs/lstm_MOA-close.pdf}
        \vspace{-0.7cm}
        \caption{location \textit{A}}
        \end{subfigure}
    \end{minipage}
    \begin{minipage}[t]{0.47\columnwidth}
        \begin{subfigure}[b]{\textwidth}
           \includegraphics[width=\linewidth]{figs/lstm_MOA-far.pdf}
        \vspace{-0.7cm}
        \caption{location \textit{B}}
         \end{subfigure}
    \end{minipage}
    \vspace{-0.4cm}
    \caption{Topological passing classification and its probability as per proposed intention-awareness action and corresponding inference.}
    \label{fig:lstm-output}
    \vspace{-0.3cm}
\end{figure} 
}

\subsection{Real Marine Accident Case Study} \label{sec:case-study}

\begin{figure}[t!]
\begin{center}
\begin{tabularx}{\columnwidth}{*{2}{>{\centering\arraybackslash}X}}
   \centering
    \includegraphics[width=0.4\columnwidth,valign=m]{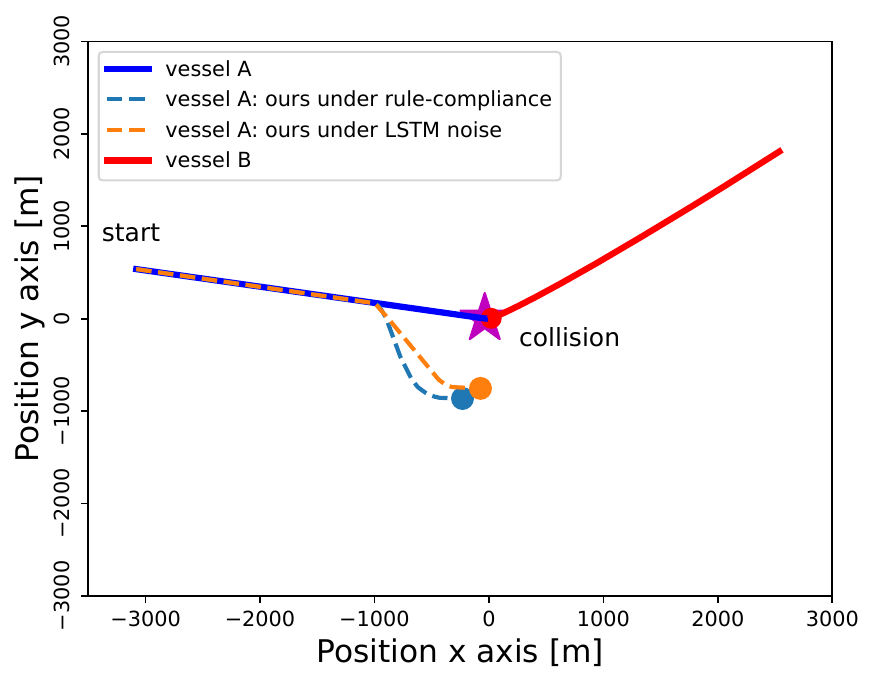} 
    &
    \centering
    \resizebox{0.35\columnwidth}{!}{%
        \begin{tabular}{c|c|c}
            \hline
            \textbf{Vessel} & \textbf{A} & \textbf{B} \\ \hline
            Length [\si{\meter}] & 12.6 & 225.0 \\ \hline
            Beam [\si{\meter}] & 3.2 & 32.3 \\ \hline
            Heading [\si[detect-weight=true]{\degree}] & 100 & 225\\ \hline
            Speed [\si[detect-weight=true]{\meter/\second}] & 5.8 & 5.7\\ \hline
        \end{tabular}
    }
\end{tabularx}
   \newline
    \centering
    \begin{minipage}[t]{0.3\columnwidth}
        \begin{subfigure}[b]{\textwidth}
           \includegraphics[width=\linewidth]{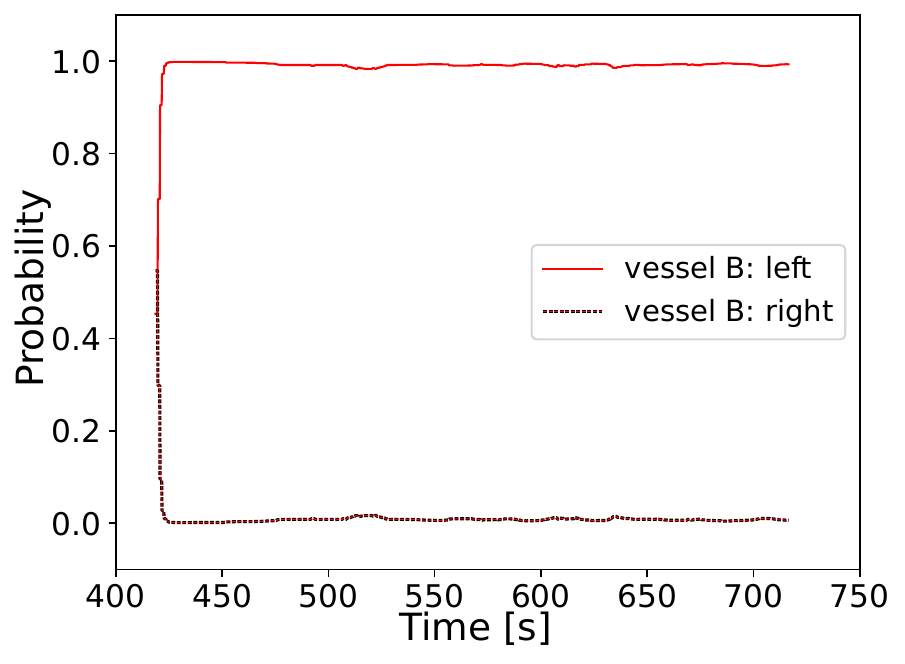}
           \vspace{-0.65cm}
            \caption{}
            \label{fig:accident-rule-compliance}
         \end{subfigure}
    \end{minipage}
    \begin{minipage}[t]{0.3\columnwidth}
        \begin{subfigure}[b]{\textwidth}
           \includegraphics[width=\linewidth]{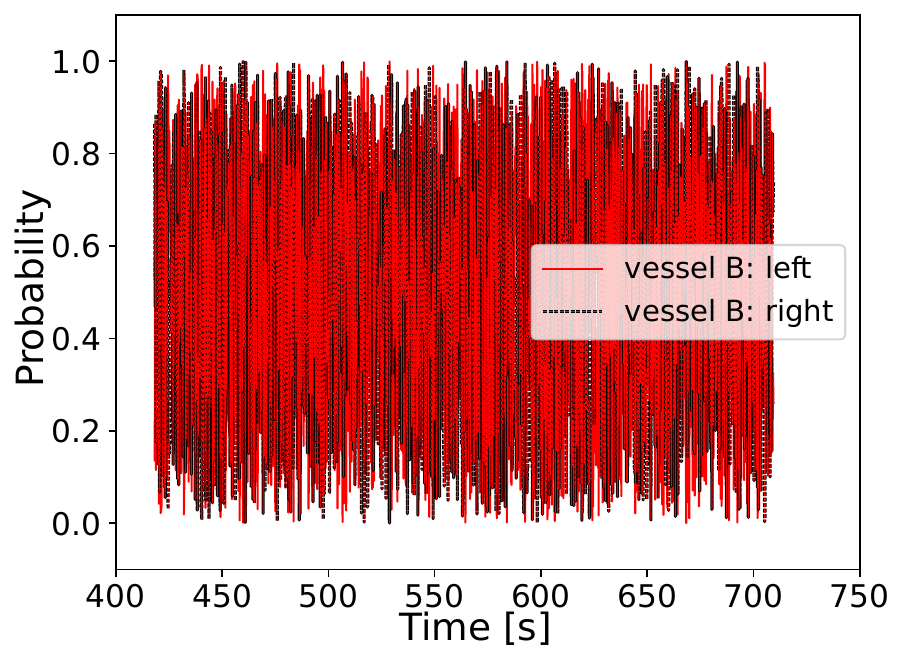}
           \vspace{-0.65cm}
           \caption{}
           \label{fig:accident-lstm-sensitivity}
         \end{subfigure}
    \end{minipage}
    \begin{minipage}[t]{0.33\columnwidth}
        \begin{subfigure}[b]{\textwidth}
           \includegraphics[width=\linewidth]{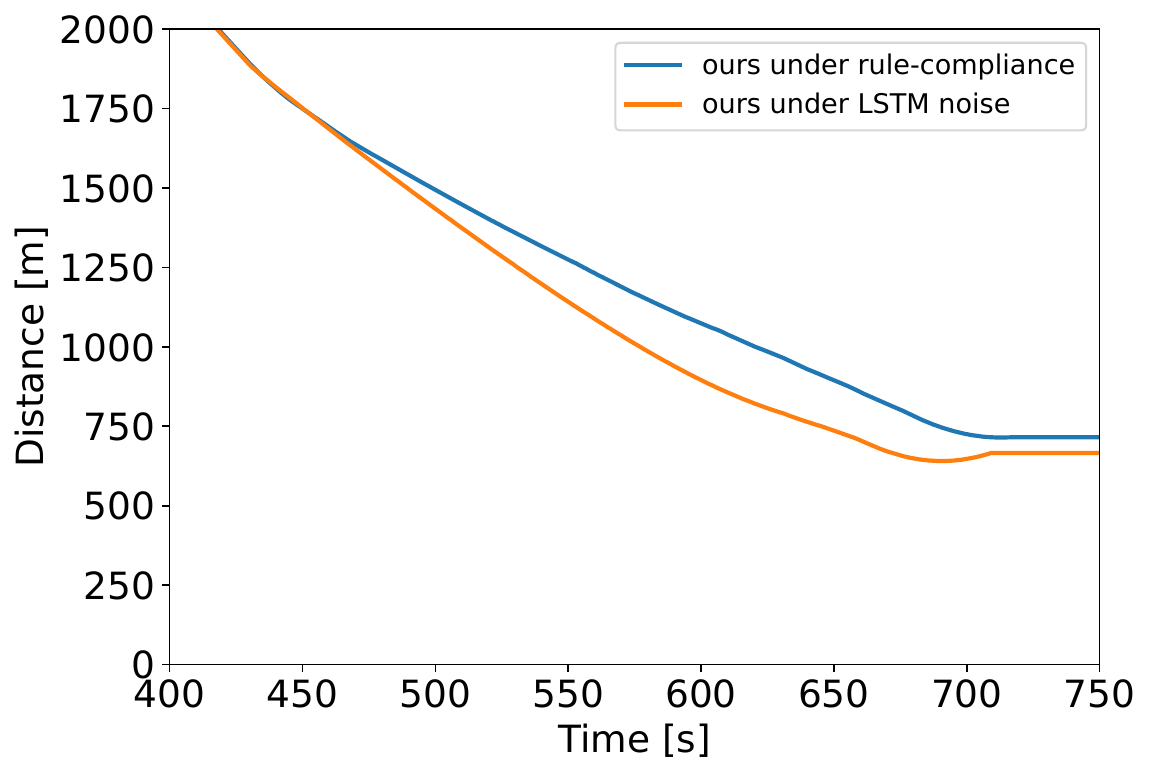}
           \vspace{-0.65cm}
           \caption{}
           \label{fig:accident-distance-compare}
         \end{subfigure}
    \end{minipage}
 \end{center}
\vspace{-0.6cm}
\caption{Collision accident case study. \textit{(top)} Trajectories of vessel \textit{A, B} during the collision (\textit{solid line}) and details of vessels involved. Trajectory of vessel \textit{A} in compliance with the rule (\textit{dotted blue}) and robustness test to random LSTM prediction (\textit{dotted orange}); (\textbf{a}) passing prediction of vessel \textit{B} by vessel \textit{A} in compliance of the rule when encountering \textit{B} at approximately \SI{420}{\second}; (\textbf{b}) random passing prediction for robustness test to LSTM; and (\textbf{c}) distance between vessel \textit{A, B} during corresponding tests.}
\vspace{-0.7cm}
\label{fig:ais-accident}
\end{figure}

We tested applicability of the proposed method to (1) different-scale vehicles; (2) actions with \textit{rule compliance} preferred; and (3) robustness to LSTM prediction, in a real accident case study as shown in \fig{fig:ais-accident}. The \textit{rule compliance} corresponds to giving a weight to $\mathcal{P}_{l}$ passing progress with $(+)$ sign, which is aligned with the rules. We empirically mapped $\Tilde{I}$ with $(+)$ winding angle by $0.3$ factor, i.e., preference of $\mathcal{P}_{l}$ to $\mathcal{P}_{r}$. We utilized historical AIS records of a collision off Cape Kodomari ($\SI{41}{\degree}11'$ N, $\SI{139}{\degree}58'$ E), Japan, on 2015 \cite{JTSB_2017}. The main cause of the collision is that vessel \textit{B} did not take an evasive action as a \textit{give-way} vessel, while vessel \textit{A} did not take a best-aid action even if the vessel \textit{B} did not follow the rule, and misinterpreted \textit{B}'s intention as passing ahead of \textit{A}. On the other hand, our proposed approach predicts the passing intention of vessel \textit{B} well (\fig{fig:accident-rule-compliance}), i.e., $\mathcal{P}_{l}$, and safely avoids \textit{B} by \textit{rule-compliance}. We also tested the robustness to LSTM incorrect predictions (\fig{fig:accident-lstm-sensitivity}), showing that, even if slightly closer to the obstacle than the case with correct LSTM prediction, the ASV is able to safely avoid the obstacle (\fig{fig:accident-distance-compare}). 
This is possible because our multi-objective optimization framework evaluates actions outside of \textit{no-go-zone} as well as prevents chattering.

\section{Conclusion and Future Steps}\label{sec:discussion}
Our proposed active, intention-aware obstacle avoidance method in multi-encounters can achieve safer navigation compared to state-of-the-art approaches. This is accomplished by introducing \textit{topological} modeling of passing based on winding numbers, \textit{passing intention classification} using an LSTM\revisedcom{-backbone neural network classifier} trained on both real-world AIS and synthetic data, and employing \revisedcom{active collision avoidance based on} multi-objective optimization \revisedcom{covering} \textit{information gain} in uncertain scenarios. %
We employ the proposed active intention-aware method, \revisedcom{validated on repetitive Monte Carlo simulations as well as a real accident case study, and} integrated into a real ASV. 

Our \revisedcom{future work} is to investigate attention-based architectures and various RNNs to effectively capture changes in the motion of obstacles. This will enable ASVs to adaptively select the motion data prediction time window and filter the samples. Furthermore, we will expand the proposed approach, for balancing rule compliance with \revisedcom{an external force- and interaction-aware planner}.

\invis{
\section*{Acknowledgement}
We would like to thank Monika Roznere and Sam Lensgraf for help with field experiments and McGill Bellairs Research Institute for experimental sites. This work is supported in part by the Burke Research Initiation Award, NSF CNS-1919647, 2144624, OIA1923004 and NH Sea Grant.
}
\printbibliography

\end{document}